\documentclass[lettersize,journal]{IEEEtran}
\usepackage{amsmath,amsfonts}
\usepackage{algorithmic}
\usepackage{algorithm}
\usepackage{array}
\usepackage{subcaption}
\usepackage{textcomp}
\usepackage{stfloats}
\usepackage{url}
\usepackage{verbatim}
\usepackage{graphicx}
\usepackage{cite}
\usepackage{hyperref}
\usepackage{multirow}
\usepackage[table,xcdraw]{xcolor}
\usepackage{placeins} % Load the placeins package for \FloatBarrier
\usepackage{cite}
\usepackage{tikz}
\usetikzlibrary{shapes.multipart}
\usetikzlibrary{shapes,arrows}
\usepackage{amsmath}

\begin{document}

\title{Deep Learning-Based Robust Multi-Object Tracking via Fusion of mmWave Radar and Camera Sensors}

\author{Lei Cheng,~\IEEEmembership{Graduate Student Member,~IEEE}, Arindam Sengupta, \\ and Siyang Cao,~\IEEEmembership{Senior Member,~IEEE}
        % <-this % stops a space
\thanks{The authors are with the Department of Electrical and Computer Engineering, The University of Arizona, Tucson, AZ 85721 USA (e-mail: leicheng@arizona.edu; sengupta@arizona.edu; caos@arizona.edu)}% <-this % stops a space
}

\markboth{Journal of \LaTeX\ Class Files,~Vol.X, No.X, X}%
{Shell \MakeLowercase{\textit{et al.}}: A Sample Article Using IEEEtran.cls for IEEE Journals}


\maketitle

\begin{abstract}

Autonomous driving holds great promise in addressing traffic safety concerns by leveraging artificial intelligence and sensor technology. Multi-Object Tracking plays a critical role in ensuring safer and more efficient navigation through complex traffic scenarios. This paper presents a novel deep learning-based method that integrates radar and camera data to enhance the accuracy and robustness of Multi-Object Tracking in autonomous driving systems. The proposed method leverages a Bi-directional Long Short-Term Memory network to incorporate long-term temporal information and improve motion prediction. An appearance feature model inspired by FaceNet is used to establish associations between objects across different frames, ensuring consistent tracking. A tri-output mechanism is employed, consisting of individual outputs for radar and camera sensors and a fusion output, to provide robustness against sensor failures and produce accurate tracking results. 
Through extensive evaluations of real-world datasets, our approach demonstrates remarkable improvements in tracking accuracy, ensuring reliable performance even in low-visibility scenarios.
\end{abstract}

\begin{IEEEkeywords}
multi-object tracking, radar, radar and camera, deep learning, sensor fusion, Bi-LSTM.
\end{IEEEkeywords}

\section{Introduction}
\IEEEPARstart{A}{utonomous} driving is gaining popularity as a solution to the pressing problem of road safety. With millions of people dying in car crashes every year (1.35 million globally \cite{who}), the potential impact of autonomous driving is vast. By leveraging deep learning and advanced sensors, autonomous driving can mitigate accidents caused by human error and optimize traffic flow for enhanced mobility \cite{tang2022road}. 
Multi-object tracking (MOT) plays a crucial role in autonomous driving. It involves distinguishing and tracking multiple objects over time, providing valuable information on motion patterns and driving behaviors of each traffic participant \cite{chiu2021probabilistic}. This enables autonomous vehicles to make informed decisions, improve motion forecasting, and avoid potential hazards. 

However, despite its importance, tracking multiple objects in unconstrained environments remains exceptionally challenging, with achieving human-level accuracy still an elusive goal \cite{milan2017online}. Currently, the majority of MOT systems rely solely on visual data, which presents several difficulties. For instance, objects may become occluded or undergo abrupt appearance changes \cite{luo2021multiple}, making them difficult to distinguish from other objects and leading to loss of track. Reflections in mirrors or windows can create confusion, and objects that closely resemble each other can lead to misidentified objects and incorrect associations \cite{leal2015motchallenge}. 

To address these limitations, researchers are exploring sensor fusion, combining data from cameras, radars, and lidars to enhance MOT systems.
Sensor fusion brings several benefits to MOT. Firstly, it ensures that the MOT system remains robust and resilient in the face of unexpected events and anomalies \cite{hackett1990multi}. When one sensor, like a visual camera, encounters issues like being suddenly blinded, other sensors like radar or lidar step in, ensuring uninterrupted tracking. Conversely, if radar or lidar faces interference, visual data can fill in the gaps to maintain tracking.
Secondly, sensor fusion improves MOT's adaptability to challenging environments and localization accuracy. Radars excel in adverse weather and low-light conditions \cite{liu2021robust}, enhancing the MOT system's robustness in such scenarios. Moreover, radar and lidar provide precise object localization \cite{kim2020extended}, which is particularly helpful in situations with object occlusions or similar appearances.

While lidar-camera fusion has shown promising results in MOT, its practical adoption has been hindered by high costs and susceptibility to adverse weather conditions \cite{hao2022asynchronous}. In contrast, radar-camera fusion provides a cost-effective alternative that fulfills all the required criteria for reliable perception \cite{barbosa2023camera}. Therefore, our focus in this paper is on radar-camera fusion to enhance the performance of MOT systems.

Deep learning is revolutionizing computer vision and natural language processing, with growing potential in MOT \cite{zhu2022looking}. Its capacity to uncover intricate, latent attributes, often elusive to humans, opens new avenues for enhancing MOT. However, most deep learning-based MOT methods focus on optical sensors like cameras and lidars, neglecting the advantages of integrating radar sensors \cite{barbosa2023camera,nabati2021centerfusion}. While some efforts combine radar and camera data, they often rely on traditional Bayesian filtering-based motion models like the Kalman filter, underutilizing deep learning's potential \cite{milan2017online,karle2023multi}. These traditional motion models tend to neglect the long-term temporal information \cite{sadeghian2017tracking} that is crucial for reliable object tracking.

\begin{figure*}[h!]
	\centering
	\includegraphics[width=0.98\textwidth]{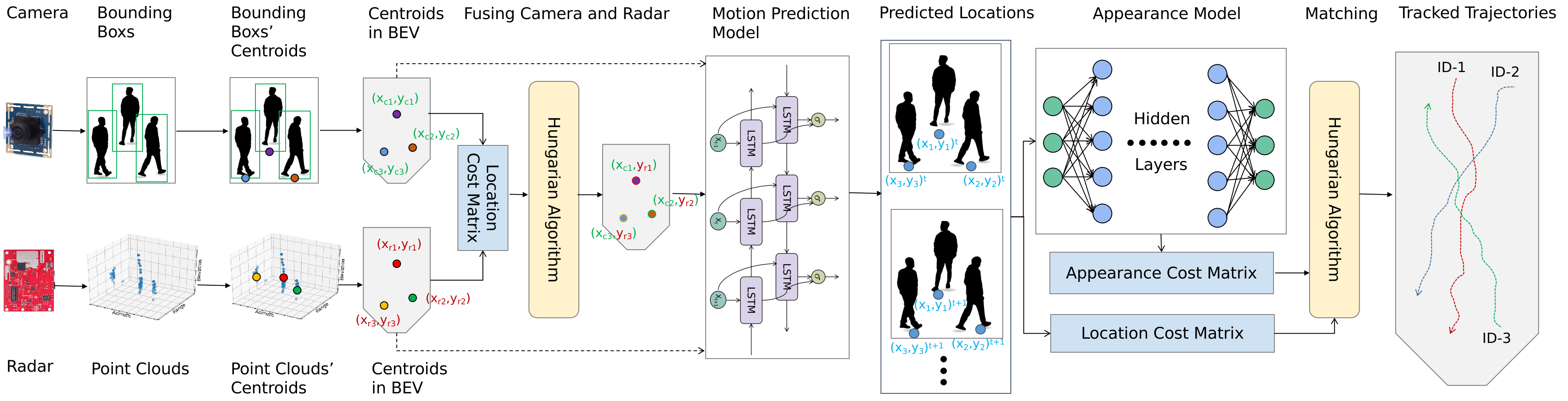}
	\caption{The general framework of the proposed MOT method. Only the sensor fusion tracker is shown here; the radar and camera trackers follow the same processing steps.}
	\label{framework}
\end{figure*}

To address this gap and build upon our previous works \cite{sengupta2022robust,sengupta2019dnn,cheng20233d,cheng2023online}, we propose a novel deep learning-based tracking method that combines radar and camera data to improve the performance of MOT systems. To achieve this, we utilize the power of deep learning models, such as Bi-LSTM (Bidirectional Long Short-Term Memory) \cite{graves2005framewise}, to incorporate long-term temporal information, enabling better motion prediction.
Additionally, we leverage the appearance feature model inspired by FaceNet \cite{schroff2015facenet} to establish associations between objects across different frames, ensuring accurate and consistent tracking.
The proposed tri-output mechanism, with individual outputs for radar and camera sensors and a fusion output, allows continuous tracking of objects and compensates for intermittent missed detections from individual sensors. This mechanism ensures that even if one of the sensors fails or experiences a temporary malfunction, the system can still operate effectively. Moreover, the fusion output merges complementary information from both radar and camera sensors, thereby enhancing tracking accuracy. As a result, our method ensures uninterrupted and reliable object tracking, significantly improving the overall robustness and performance of the MOT system, particularly in terms of enhancing tracking accuracy and reducing identity (ID) switches. The framework of the proposed method is illustrated in Fig. \ref{framework}, while a more detailed process is depicted in Fig. \ref{pipe_realtime}. The main contributions of our work can be summarized as follows:
\begin{enumerate}
  \item  We introduce a deep learning-based MOT method fusing both radar and camera data for enhanced accuracy and robustness.
  \item  We employ a FaceNet-derived appearance feature model to better associate objects across different frames, reducing ID switches.  
  \item  We utilize Bi-LSTM networks to incorporate long-term temporal information and better predict object motions.
  \item  We use a tri-output structure, with one output for each individual sensor and one for the decision-level fusion layer, to ensure robustness to sensor failures and provide accurate tracking results.  
\end{enumerate}

The rest of this paper is organized as follows: Section II presents a review of related works. Section III provides preliminary knowledge in the fields of MOT, and radar-camera sensor fusion. Section IV describes the proposed method in detail. Section V presents experimental results and analysis, followed by conclusions and future work in Section VI.

\section{Related Works}
MOT has been extensively researched in separate domains of radar and cameras. Recently, MOT based on computer vision has experienced rapid development and has become a research hotspot. However, MOT that leverages the fusion of radar and camera sensors is still relatively less explored but holds great promise for the future.
%\subsection{Visual-based Multi-Object Tracking}

Authors in \cite{bewley2016simple} presented an approach called Simple Online and Realtime Tracking (SORT), which combines a CNN-based detector with the Kalman filter and Hungarian algorithm, for efficient online tracking. Authors in \cite{wojke2017simple} proposed Deep SORT to reduce identity switches by integrating the appearance information learned through a CNN model trained on a person re-identification (ReID) dataset. This work highlighted the significance of appearance information in multi-object tracking (MOT). Many studies have thus focused on efficiently exploring distinctive appearance features, particularly through deep metric learning and local feature learning methods, as introduced in \cite{ming2022deep, cheng20233d, schroff2015facenet}, leading to the emergence of the Person-ReID task. Recent MOT endeavors \cite{fischer2023qdtrack,sun2022dancetrack,wang2020towards,chen2018real} are leveraging ReID appearance features based on deep metric learning to enhance tracking performance. FaceNet \cite{schroff2015facenet} stands out as a pioneering application of deep metric learning, designed to automatically learn discriminative ReID features from images and produce distance metric feature vectors \cite{ge2018deep}. It aligns perfectly with our MOT needs, leading us to adopt FaceNet as our appearance model.
Authors in \cite{milan2017online} introduced a pioneering recurrent neural network (RNN) framework for MOT. This approach demonstrates RNN's capability of learning complex motion models in realistic environments. Subsequently, works in \cite{xue2020poppl,yang2022ais} advanced this by incorporating Bi-LSTM for object tracking, achieving improved accuracy in trajectory prediction. Authors in \cite{zhu2022looking} propose MO3TR, a transformer-based framework for MOT that leverages long-term temporal information to handle occlusions and track initiation/termination without explicit data association or heuristics. These studies have provided insights into the potential benefits of integrating multiple cues to enhance MOT performance, such as employing CNNs to learn discriminative ReID appearance features and utilizing RNNs or transformers to capture long-term real-world motion patterns with noise and nonlinearity. However, it is important to note that all of these efforts have been solely based on visual information and not on the fusion of radar and camera data. As mentioned earlier, visual-based MOT faces challenges that cannot be fully addressed on its own.
Therefore, there is a need to explore the integration of radar and vision data to improve MOT performance.

Authors in \cite{hao2022asynchronous} proposed a fusion estimator for MOT by integrating the advantages of a camera and a radar. Camera and radar data are fused using a matrix-weighted approach based on matching results through the Hungarian algorithm.
Authors in \cite{liu2021robust} proposed a novel approach to MOT by fusing radar and camera information to address the limitations of single sensors in severe weather conditions. They applied the Kalman filter-based Joint Probabilistic Data Association (JPDA) method for decision-level radar and camera data fusion to improve tracking robustness. 
Authors in \cite{karle2023multi} proposed the first real-world implementation of a radar-camera late fusion and tracking algorithm for high-speed racing, based on an Extended Kalman Filter (EKF). This approach is predicated on the belief that late decision-level fusion can fuse heterogeneous, multi-modal detection input to generate an optimized, unique object list. 
Authors in \cite{cui2023online} introduced a fused radar-camera MOT system, where fusion is accomplished through a graph-based association method that simultaneously considers motion similarity and appearance similarity. They employ a metric learning network to extract distinctive deep features and utilize the EKF for motion state estimation.
These works focus on enhancing MOT performance through the fusion of radar and camera sensors. However, they either fail to explore the potential of harnessing deep learning or only utilize deep learning for extracting appearance features. Some other radar-camera fusion works focus on 3D MOT \cite{li2023poly}. For example, \cite{zhou20223d} employs an optimized spatial-temporal unification of radar-camera information along with a self-adaptive data association method, which leverages motion, appearance, and geometry features comprehensively, to improve 3D MOT. \cite{kim2023crn} introduces the Camera Radar Net to generate a semantically rich and spatially accurate BEV (Bird's Eye View) feature map by fusing the complementary characteristics of camera and radar sensors, thereby enhancing 3D MOT. However, we endeavor to implement a radar-camera fusion-based 2D MOT in a road plane or BEV.
Based on the above discussion, we propose a novel 2D MOT approach that involves using FaceNet-inspired deep metric learning to extract discriminable ReID appearance features, Bi-LSTM for improved motion trajectory prediction, and a tri-output mechanism to account for sensor failures. The integration of these components aims to enhance the accuracy and robustness of multi-sensor fusion in MOT, making it a significant advancement in the field.

\section{Preliminaries}
\subsection{Multi-Object Tracking}
%Multi-object tracking has gained growing interest in recent years, owing to its far-reaching potential for various fields of research and practical applications. 
MOT can be represented as follows: Let $\mathcal{F}$ be the set of frames in a data sequence, where $\mathcal{F} = \{F_1, F_2, ..., F_T\}$ and $T$ is the total number of frames. In each frame $F_t$, there exist multiple object instances represented by bounding boxes or clusters denoted as $\mathcal{O}_t = \{O_{t,1}, O_{t,2}, ..., O_{t, n_t}\}$, where $n_t$ is the number of object instances in frame $t$. The goal is to establish associations across frames to create tracks $\mathcal{T} = \{T_1, T_2, ..., T_K\}$ accurately estimating the trajectories of multiple objects and re-identifying them across consecutive frames captured by one or multiple sensors, where $K$ is the total number of tracks.  
Developing MOT approaches involves tackling three critical steps: measuring similarity between objects in frames, recovering identity information based on similarity measurements \cite{luo2021multiple}, and managing track life cycles. 

\subsubsection{Similarity computation} 
Motion and appearance are two essential cues for computing the similarity between detections and tracks in MOT. The motion cues-based method uses motion models to predict the potential position of tracks in future frames, thereby reducing the search space \cite{sadeghian2017tracking}. It can be expressed as:
\begin{equation}
\mathbf{t}_{j,t+1} = \mathbf{M}(\mathbf{t}_{j,t}) ,
\end{equation}
where $\mathbf{M}$ is the motion model function, $t$ denotes the current frame index, and $\mathbf{t}_{j,t}$ is the track corresponding to object $j$ at frame $t$. Linear motion models, such as the Kalman Filter, are a simple yet effective preliminary choice for motion models but struggle with real-world motion changes, where non-linear models can alleviate such issue to some extent \cite{sun2019deep}.

The appearance cues-based method employs sophisticated models of object appearance, such as deep neural networks, to differentiate between objects by extracting distinctive features \cite{wojke2017simple}. These features are then utilized to match tracks and rediscover lost objects, significantly reducing identity switches and improving tracking accuracy.
The similarity between a detection $O_{t,i}$ in frame $F_t$ and a track $T_k$ can be determined using a similarity function $s(O_{t,i}, T_k)$, often computed using a distance metric like the Euclidean distance. This function will compute both appearance and motion cues through separate similarity measures: $s_{\text{appearance}}(O_{t,i}, T_k)$ for appearance and $s_{\text{motion}}(O_{t,i}, T_k)$ for motion. The overall similarity can be represented as:
\begin{equation}
s(O_{t,i}, T_k) = \alpha \cdot s_{\text{appearance}}(O_{t,i}, T_k) + \beta \cdot s_{\text{motion}}(O_{t,i}, T_k),
\end{equation}
here, $\alpha$ and $\beta$ are weighting factors controlling the appearance and motion cues' contribution to the overall similarity score.

\subsubsection{Association strategy} 
The association stage links detections across frames to the same object, despite appearance changes or detection gaps. It inputs new detections and existing trajectories, aiming for a consistent relationship between them by evaluating all possible detection-trajectory combinations to ensure optimal one-to-one matching \cite{zhang2020long, luo2021multiple}.

Therefore, the association between detections and tracks can be formulated as an optimization problem. Let $x_{t,i} \in \{0, 1\}$ denote whether detection $O_{t,i}$ is associated with any track in frame $t$, and $y_{t,j} \in \{0, 1\}$ denote whether track $T_j$ is active in frame $t$. The association matrix can be represented as $\mathbf{X} = [x_{t,i}]$ and $\mathbf{Y} = [y_{t,j}]$. The optimization objective is to minimize the sum of similarity scores while ensuring each detection is optimally associated with only one track:
\begin{equation}
\begin{aligned}
\min_{\mathbf{X}, \mathbf{Y}} \quad & \sum_{t=1}^{T} \sum_{i=1}^{n_t} \sum_{j=1}^{K} s(O_{t,i}, T_j) x_{t,i} y_{t,j}, \\
\quad \text{subject to} \\ \quad & \sum_{j=1}^{K} y_{t,j} = 1, \quad \forall t = 1, 2, ..., T, \\
& \sum_{i=1}^{n_t} x_{t,i} = 1, \quad \forall t = 1, 2, ..., T.
\end{aligned}
\end{equation}

To solve this problem, methods like the Hungarian algorithm or greedy assignment are often used to find the best associations\cite{baser2019fantrack}.
\subsubsection{Track management} 
%Effective track life cycle management is a critical element in ensuring the reliability and accuracy of MOT systems. This process 
Track management oversees the entire lifecycle of tracks, including initiation, state estimation, updates, and termination \cite{baser2019fantrack,chu2019famnet}. Mathematically, they can be represented as follows:

Initiation: When a new detection $z_\mathbf{t}^l$ is made at time step $l$, a new track $\mathbf{t}$ is initiated with an initial state $\hat{s}_\mathbf{t}^l$ determined using the motion model $f(\cdot)$:
\begin{equation}
\hat{s}_\mathbf{t}^l  = f(z_\mathbf{t}^l)
\end{equation}

State estimation: At time step $l$, the state of track $\mathbf{t}$ is estimated as $\hat{s}_\mathbf{t}^l$, based on the previous state $\hat{s}_\mathbf{t}^{l-1}$, the measurement $z_\mathbf{t}^l$, and the motion model $f(\cdot)$:
\begin{equation}
\hat{s}_\mathbf{t}^l = f(\hat{s}_\mathbf{t}^{l-1}, z_\mathbf{t}^l)
\end{equation}

Updates: The state of track $\mathbf{t}$ is updated at each time step $l$ based on the estimated current state $\hat{s}_\mathbf{t}^{l}$, the current measurement $z_\mathbf{t}^l$ and the motion model $f(\cdot)$:
\begin{equation}
\hat{s}_\mathbf{t}^{l+1} = f(\hat{s}_\mathbf{t}^l, z_\mathbf{t}^l)
\end{equation}

Termination: When a track $\mathbf{t}$ has not been updated (maintained the same state) for a certain number of time steps $L$, it is terminated:
%$$\hat{x}_\mathbf{t}^l \neq \emptyset, \quad l < T \Rightarrow \hat{x}_\mathbf{t}^l = \emptyset$$
\begin{equation}
\hat{s}_\mathbf{t}^{l-L+1} = \hat{s}_\mathbf{t}^{l-L+2} = \ldots = \hat{s}_\mathbf{t}^l \neq \emptyset \quad \Rightarrow \quad 
 \hat{s}_\mathbf{t}^l = \emptyset
\end{equation}
% \[
% \text{If } \hat{x}_\mathbf{t}^{l-T+1} = \hat{x}_\mathbf{t}^{l-T+2} = \ldots = \hat{x}_\mathbf{t}^l \neq \emptyset, \text{ then } \hat{x}_\mathbf{t}^l = \emptyset
% \]
Here, $\hat{s}_\mathbf{t}^l$ represents the state of track $\mathbf{t}$ at time step $l$, $z_\mathbf{t}^l$ represents the measurement of track $\mathbf{t}$ at time step $l$, $f(\cdot)$ represents the motion or dynamics model, and $L$ represents the number of time steps after which a track is considered terminated.
Importantly, track management dictates when to initiate and terminate each track, a decision that carries considerable weight in minimizing false positives and identity switches. Incorrect timing can result in excessive tracks, increased delays, and reduced accuracy. Thus, tuning the parameters controlling track initiation and termination is essential.

\subsection{Radar-Camera Sensor Fusion}
Multi-sensor fusion, a technique that merges information from multiple sensors for unique inferences not possible with a single sensor, involves steps such as gathering, preprocessing, comparing, and combining sensor readings for the final result \cite{liggins2017handbook}. Specifically, in MOT systems, fusing radar point clouds and camera data provides a comprehensive and accurate environmental perception, enabling robust tracking \cite{liu2021robust, wang2019multi}.

\subsubsection{Radar-Camera Co-Calibration}
%The sensor fusion process can be decomposed into several steps: gathering, preprocessing, comparing, and combining sensor readings to obtain the final result. 
A crucial preprocessing step for comparing data from independent physical sensors in sensor fusion is the transformation of all input data into a common coordinate system, known as calibration \cite{liggins2017handbook}. 
Calibration is a crucial step for successful sensor fusion and typically involves two types: intrinsic and extrinsic calibration \cite{pervsic2021spatiotemporal}.

Intrinsic calibration focuses on estimating the internal parameters specific to each sensor \cite{domhof2021joint,ikram2019automated}. 

Extrinsic calibration, also known as spatial calibration, involves estimating the spatial transformation between the sensor coordinates and other sensors or unified reference frames \cite{domhof2019extrinsic}. Radar-camera co-calibration involves solving for a transformation matrix that establishes a correspondence between a point, denoted as $P_p=(u,v)$, in the image pixel coordinate system (PCS) and another point, denoted as $P_r=(X_r,Y_r,Z_r)$, in the radar coordinate system (RCS) \cite{cheng20233d,cheng2023online}. It can be represented as:
\begin{equation}\label{eq1}
\small
s\begin{bmatrix}
  u  \\
  v  \\
  %\vdots \\
  1
\end{bmatrix}  
=
%\overset{\mathbb{K}}{\overbrace{\begin{bmatrix}
\underset{\mathbb{K}}{\underbrace{\begin{bmatrix}
  f_x & \gamma & u_0 \\
  0 & f_y & v_0 \\
  0 & 0 & 1 \\
\end{bmatrix}}}
\underset{\mathbb{R|T}}{\underbrace{\begin{bmatrix}
  r_{11} & r_{12} & r_{13} & t_x \\
  r_{21} & r_{22} & r_{23} & t_y \\
  r_{31} & r_{32} & r_{33} & t_z \\
\end{bmatrix}}}
\begin{bmatrix}
  X_r  \\
  Y_r  \\
  Z_r  \\
  1
\end{bmatrix} 
\end{equation}
where $f_x$ and $f_y$ are the camera focal lengths in pixel units, $(u_0,v_0)$ represent the principal point, $s$ is the scaling factor and $\gamma$ is the skew coefficient between the image axes, $r_{ij}$ are the rotation matrix elements, $t_x$, $t_y$, and $t_z$ are the translation parameters.
This transformation matrix comprises the intrinsic matrix $K$ and the extrinsic matrix $[R|T]$, obtained through intrinsic calibration and extrinsic calibration, respectively. 

\subsubsection{Sensor Fusion Schemes}
Various sensor fusion schemes have been developed to effectively combine multiple sensing modalities. These schemes include data-level fusion, feature-level fusion, and decision-level fusion.
In data-level fusion, raw or pre-processed sensory data from different sensor modalities are fused together. Deep learning techniques are commonly used in this level of fusion to learn a joint representation from the sensing modalities \cite{nabati2021centerfusion}. 
Feature-level fusion focuses on extracting representative features from sensor data. In the case of multi-sensor feature-level fusion, features are extracted from multiple sensor observations and combined into a single concatenated feature vector that serves as input to pattern recognition techniques such as neural networks, clustering algorithms, or template methods  \cite{liggins2017handbook}. 
Decision-level fusion, on the other hand, combines sensor information at the decision-making stage, after each sensor has made a preliminary determination of an entity's location, attributes, and identity \cite{liggins2017handbook}. 
Both data-level fusion and feature-level fusion are susceptible to errors stemming from noise, sensor outliers, and spatial or temporal misalignment of the data \cite{karle2023multi,malawade2022hydrafusion,nabati2021centerfusion}, which can significantly diminish the overall robustness of these approaches. In contrast, decision-level fusion approaches may sacrifice some rich intermediate features, but they provide enhanced robustness and versatility, facilitating the seamless integration of new sensors into the multi-sensor system \cite{nabati2021centerfusion}. Therefore, our method adopts decision-level fusion, ensuring the reliability and adaptability of the sensor fusion process. 

\subsection{FaceNet and Bi-LSTM}
\subsubsection{FaceNet}
FaceNet is a deep learning model focused on face verification, which learns a mapping: $F: \mathcal{X} \to \mathbb{R}^d$, where $\mathcal{X} = \left \{\mathbf{x}_1, \mathbf{x}_2, ..., \mathbf{x}_N \right \}$ is the set of $N$ face images and $\mathbb{R}^d$ is the $d$-dimensional embedding space. It maps each face image $\mathbf{x}_i \in \mathcal{X}$ to a compact and robust face feature embeddings $\mathbf{r}_i \in \mathbb{R}^d$, such that faces with similar appearances are mapped to nearby points in the embedding space. FaceNet combines a base CNN, like Inception or MobileNets, for extracting features from faces, with a triplet loss function that ensures discriminative learning of features for accurate identity verification.

The triplet loss function operates with three images: an anchor $x_a$, a positive $x_p$ with the same identity, and a negative $x_n$ with a different one. It aims to minimize the distance between $x_a$ and $x_p$ while maximizing that between $x_a$ and $x_n$ \cite{ye2021deep}, using the formula:

\begin{equation}
\begin{aligned}
\mathcal{L}_{\text{triplet}}(x_a, x_p, x_n)
= \max(0, \alpha + d(f(x_a), f(x_n))  \\ - d(f(x_a), f(x_p))),
\end{aligned}
\end{equation}
where $d(\cdot, \cdot)$ measures embedding similarity, $f(x)$ represents the embedding generated by the base CNN for image $x$, and $\alpha$ is a margin that is used to enforce a minimum distance between embeddings of dissimilar images and embeddings of similar images.
Thus, the triplet loss attempts to establish a margin between each pair of appearances from one object to all others, guiding the model to learn discriminative embedding features that ensure ample separation between different identities \cite{schroff2015facenet}.

\subsubsection{Bi-LSTM} 
RNNs, with feedback connections, maintain a hidden state $\mathbf{h}$ that integrates information from past inputs, updating it with new inputs \cite{ma2020particle}. Among these, LSTM is a prominent RNN variant, featuring a memory cell and three gates—forget $\mathbf{f}$, input $\mathbf{i}$, and output $\mathbf{o}$—that manage information flow, enabling selective retention or omission of past states \cite{emmert2020introductory}. This architecture facilitates learning long-term dependencies in sequences.
An LSTM unit can be represented as follows:
\begin{equation}
\begin{aligned}
\quad \mathbf{i}_t & = \sigma(\mathbf{W}_i \mathbf{x}_t + \mathbf{U}_i \mathbf{h}_{t-1} + \mathbf{V}_i \mathbf{c}_{t-1} + \mathbf{b}_i) \\
\quad \mathbf{f}_t & = \sigma(\mathbf{W}_f \mathbf{x}_t + \mathbf{U}_f \mathbf{h}_{t-1} + \mathbf{V}_f \mathbf{c}_{t-1} + \mathbf{b}_f) \\
\quad \mathbf{c}_t & = \mathbf{f}_t \odot \mathbf{c}_{t-1} + \mathbf{i}_t \odot \tanh(\mathbf{W}_c \mathbf{x}_t + \mathbf{U}_c \mathbf{h}_{t-1} + \mathbf{b}_c)  \\
\quad \mathbf{o}_t & = \sigma(\mathbf{W}_o \mathbf{x}_t + \mathbf{U}_o \mathbf{h}_{t-1} + \mathbf{V}_o \mathbf{c}_{t} + \mathbf{b}_o) \\
\quad \mathbf{h}_t & = \mathbf{o}_t \odot \tanh(\mathbf{c}_t)
\end{aligned}
\end{equation}
where $\mathbf{c}$ is the cell state, $W$, $U$, $V$, and $b$ are learned weight matrices and bias vector parameters, $\sigma$ denotes the sigmoid activation function, $\tanh$ denotes the tangent activation function, and $\odot$ represents element-wise multiplication.

Unlike traditional LSTMs, which only consider the previous context, Bi-LSTMs take advantage of both the previous and future context by processing the data from opposite directions \cite{xue2020poppl} with two separate LSTM layers. The forward layer, $\text{LSTM}^{F}$, processes the input sequence from left to right, while the backward layer, $\text{LSTM}^{B}$, processes it inversely. The output of each LSTM layer is then combined, typically through concatenation, to produce the final output.
The Bi-LSTM network can be represented as follows:
\begin{equation}
\begin{aligned}
\mathbf{h}^{F}_t &= \text{LSTM}^{F}(\mathbf{x_t}; \mathbf{W}^{F}, \mathbf{U}^{F}, \mathbf{V}^{F})
\\
\mathbf{h}^{B}_t &= \text{LSTM}^{B}(\mathbf{x_t}; \mathbf{W}^{B}, \mathbf{U}^{B}, \mathbf{V}^{B})
\\
\mathbf{h_t} &= [\ \mathbf{h}^{F}_t; \mathbf{h}^{B}_t \ ]
\end{aligned}
\end{equation}
where $\mathbf{h}^{F}$ and $\mathbf{h}^{B}$ are the outputs of the forward and backward LSTM layers, respectively, $\mathbf{h}$ is the concatenation of $\mathbf{h}^{F}$ and $\mathbf{h}^{B}$. This structure lets Bi-LSTMs capture the full scope of sequential relationships in the data, considering points both before and after the current one for precise predictions\cite{graves2005framewise}.

\section{Methodology}
The proposed MOT method, based on deep learning and radar-camera sensor fusion, involves four main steps.
The first step is to utilize camera data to train a dedicated deep learning model specifically designed for learning discriminative appearance features of objects. This model will be used to identify objects and distinguish them from others in the scene. 
Next, two separate motion prediction models are trained using location data from the camera and radar, respectively. These models predict the future positions of objects based on their past movements, allowing the algorithm to anticipate potential occlusions and motion patterns.
The third step involves integrating the appearance model and the two motion prediction models to achieve effective MOT. During the association phase, we introduce a dual-cue association strategy that simultaneously considers the distance between appearance features and the distance between object positions. Then, the tracks of objects are updated based on the associated detections, and the updated tracks are used to improve the accuracy of the motion predictions. Finally, each sensor produces its individual tracking output, while a fused tracking output is generated by combining the outputs from both sensors.

\subsection{Appearance Model based on FaceNet}
In MOT, the problem of object ID switching or re-identification has always been one of the most important factors affecting its performance. Analogous to people using appearance to distinguish objects, naturally, we can also use appearance features to better identify objects. By using a deep learning model to extract the appearance features, we can compute the distance between two appearance features to know how similar these two objects are. Currently, deep metric learning, which uses neural networks to automatically learn discriminative features from images and then outputs distance metric feature vectors, is becoming a powerful method for evaluating similarity or dissimilarity between images \cite{ge2018deep}. FaceNet is a pioneer in applying deep metric learning. 

\begin{figure}[h!]
	\centering
	\includegraphics[width=0.45\textwidth]{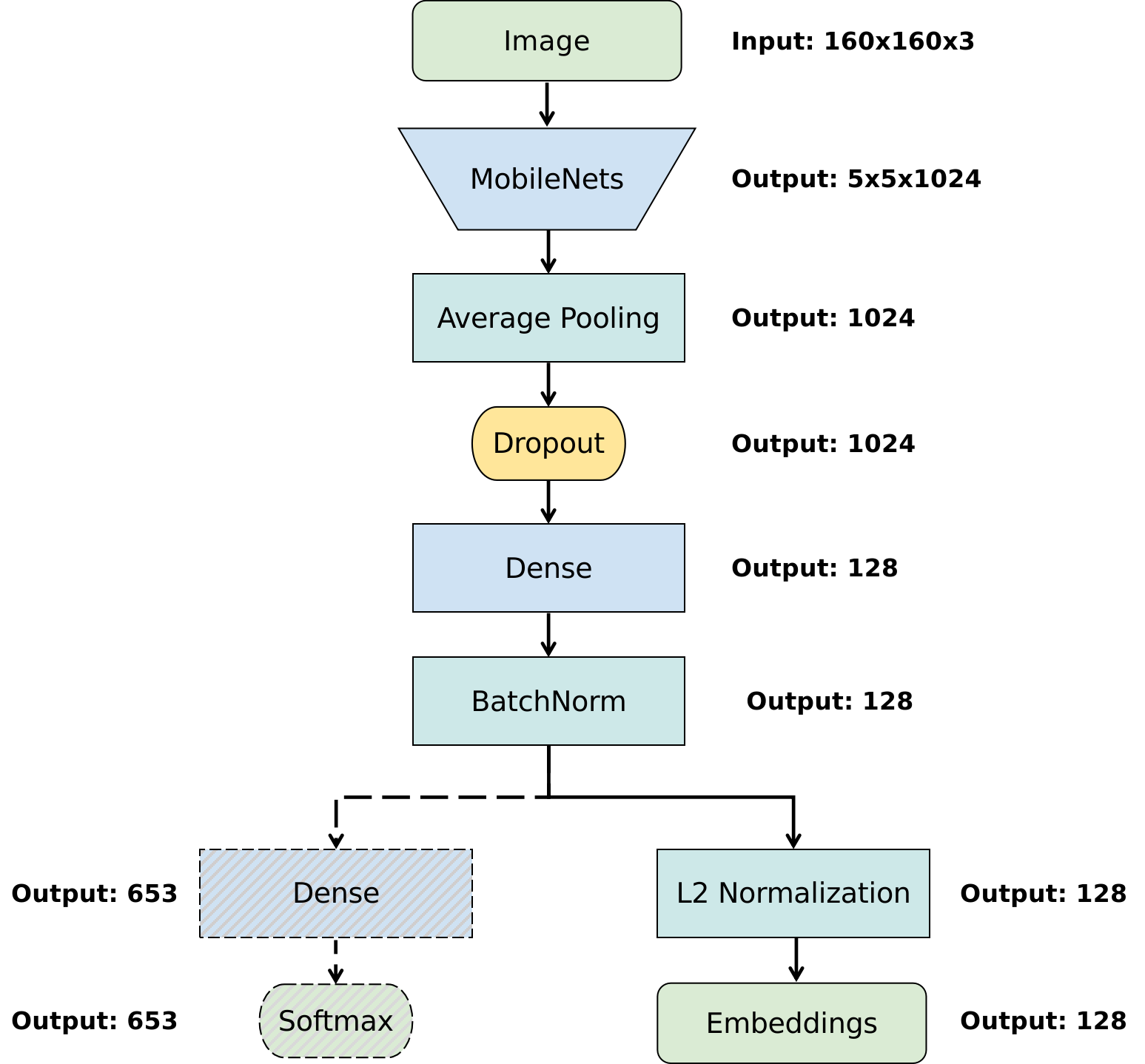}
	\caption{The network architecture of the appearance model.}
	\label{appear_m}
\end{figure}
Drawing on the success of FaceNet, we developed our own deep learning model for object appearance feature extraction to evaluate the similarity between two objects. The graph of the proposed networks is shown in Fig. \ref{appear_m}. For this purpose, we opted for the MobileNets \cite{howard2017mobilenets} as our backbone due to its fast inference speed, modest computational resource requirements, and acceptable accuracy, making it an ideal choice for MOT systems that often necessitate real-time performance and extensive computations. The output of MobileNets undergoes L2 normalization, resulting in a 128-dimensional face embedding feature, which serves as the fundamental representation in our model. Subsequently, we employ a triplet-based loss function, culminating in the completion of our model. Because training such a large and complicated model usually experiences a slower convergence, we also implement a classification branch network to aid in convergence. Note that this classification branch is only used for training and will be stripped in actual deployment, leaving the primary model intact. For this purpose, an additional softmax loss function is used in conjunction with the triplet loss.
Let $y_i$ be the one-hot encoded label for the $i$-th image, where $y_i \in \{0, 1\}^C$ and $C$ are the total number of classes (identity labels). Each element of $y_i$ is either 0 or 1, indicating the absence or presence of the corresponding class label, respectively. The softmax loss can be defined as follows:
\begin{equation}
\mathcal{L}_{\text{softmax}}(x_i, y_i) = - \log\left(\frac{e^{f{y_i}(x_i)}}{\sum_{j=1}^{C}e^{f_j(x_i)}}\right),
\end{equation}
where $f_j(x_i)$ represents the $j$-th element of the embedding $f(x_i)$ for the $i$-th image.
The overall loss function used in the model combines the triplet loss and the softmax loss:
\begin{equation}
\begin{aligned}
\mathcal{L}_{\text{total}}(x_a, x_p, x_n, y_a, y_p, y_n) = \mathcal{L}_{\text{triplet}}(x_a, x_p, x_n) + \\
\lambda (\mathcal{L}_{\text{softmax}}(x_a, y_a) + \\ \mathcal{L}_{\text{softmax}}(x_p, y_p) + \\ \mathcal{L}_{\text{softmax}}(x_n, y_n)),
\end{aligned}
\end{equation}
where $\lambda$ is a hyperparameter that controls the trade-off between the triplet loss and the softmax loss. The training objective of the model is to minimize the overall loss function $\mathcal{L}_{\text{total}}$ over the entire training dataset. This can be represented as:
\begin{equation}
\min_{\theta} \sum_{i=1}^{N} \mathcal{L}_{\text{total}}(x_a^i, x_p^i, x_n^i, y_a^i, y_p^i, y_n^i),
\end{equation}
where $N$ is the total number of training samples, and $\theta$ represents the parameters of the network.

During the training process, we implement a random sampling approach to generate triplets. Specifically, randomly select two images from the images of the same identity as an anchor and a positive, respectively, and then randomly select an image from images of different identities as a negative. Such three images - the anchor, positive, and negative - constitute a single triplet and serve as a training sample.
To efficiently train our model, we leverage the CASIA-WebFace dataset for pre-training. This dataset is a comprehensive collection of labeled faces, containing 494,414 face images from 10,575 distinct real identities, all sourced from the web \cite{yi2014learning}. Utilizing this extensive dataset for pre-training has two advantages: it helps prevent slow convergence due to random initialization and, to some extent, improves the model's ability to generalize. However, it's crucial to emphasize that our primary goal isn't facial recognition but rather distinguishing between different objects.

\begin{figure}[h!]
	\centering
	\includegraphics[width=0.35\textwidth]{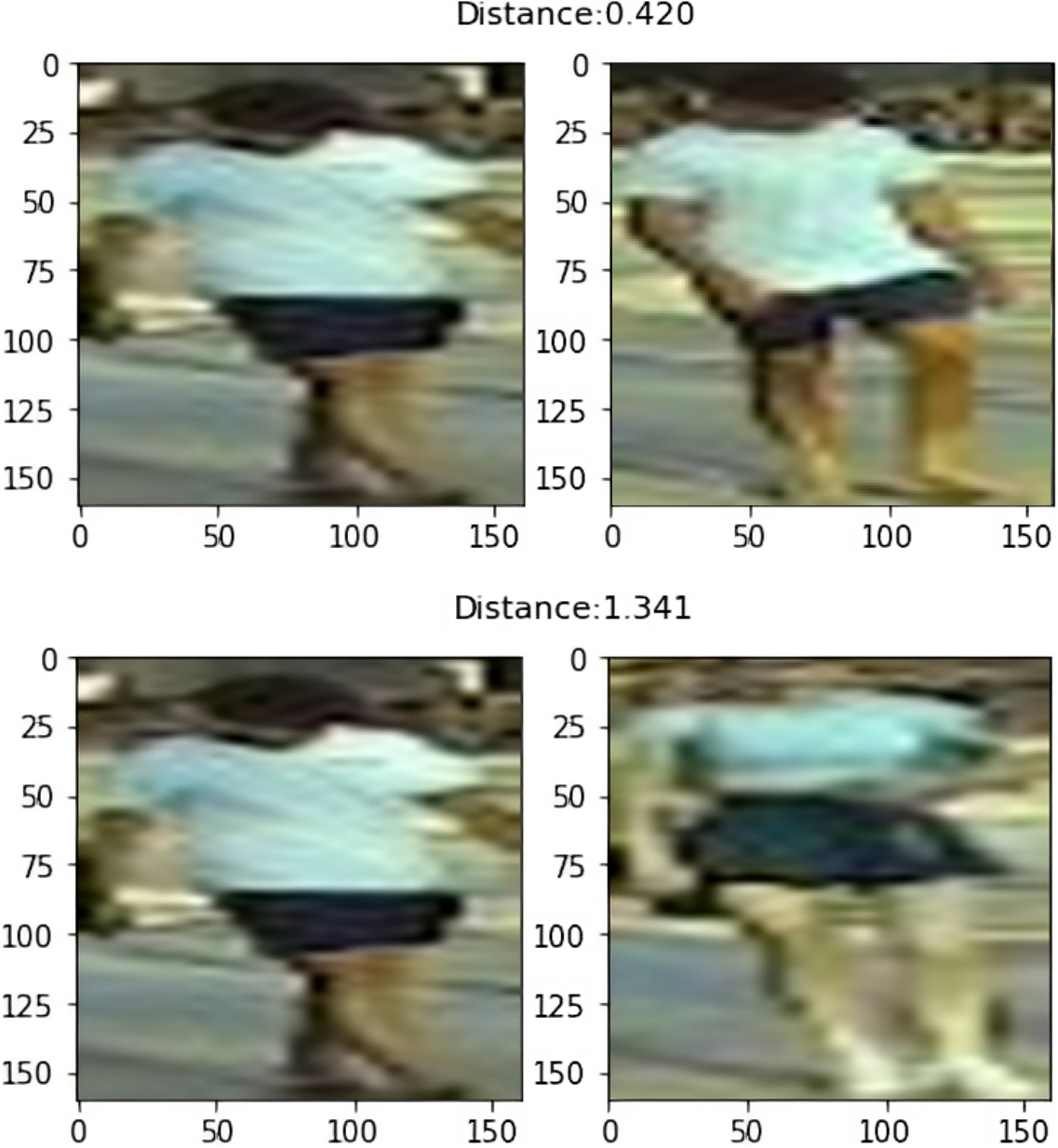}
	\caption{The distance between the same object (Above) and the distance between the different objects (Below).}
	\label{simil}
\end{figure}

Therefore, after completing the pre-training stage, we proceeded to fine-tune the same model using our own custom-collected dataset. This dataset comprises 17,994 high-quality images encompassing 653 unique identities, including both humans and vehicles. We divide the dataset into three subsets: a training set with 15,168 images, a validation set with 1,687 images, and a test set with 1,139 images. By initializing the re-training process with the pre-trained weights obtained from the CASIA-WebFace pre-training, we enable the model to benefit from the knowledge gained from the extensive facial dataset and fine-tune it for optimal performance in the context of MOT, where distinguishing between different objects is of utmost importance.
Once the model has learned the embedding space, evaluating object similarity becomes a straightforward process, involving the simple thresholding of the distance between two embedding features. Through experimentation on our testing dataset, we have determined that the optimal threshold is 1.09, resulting in an impressive classification accuracy of 98.65\%. A compelling demonstration of the appearance model's efficacy is presented in Fig. \ref{simil}, where it is evident that the distance between instances of the same identity is markedly greater than the distance between instances of different identities. 

\subsection{Motion Prediction Model based on Bi-LSTM}
After developing the appearance model, our focus shifts to implementing the motion prediction model. This model aims to forecast an object's future location by analyzing its previous locations, essentially a sequential processing task.
Unlike the Kalman Filter, which requires accurate knowledge of the underlying state-space (SS) model for optimal performance \cite{revach2022kalmannet}, our motion prediction model leverages the strengths of RNNs, specifically Bi-LSTMs, to handle the complexity of MOT. The Kalman Filter's reliance on a precise state-space model can be limiting, as obtaining accurate model parameters can be difficult, and the model may only offer a rough approximation of the complex dynamics. This becomes particularly evident in MOT tasks, where movements by vehicles and pedestrians are often unpredictable and nonlinear. Bi-LSTMs, in contrast, capture deeper temporal dependencies within sequences, crucial for accurate motion prediction in MOT. They outperform traditional LSTMs and other RNN variants in various sequence prediction tasks \cite{chen2016gentle}, mainly due to their ability to consider both past and future contexts simultaneously. In the MOT context, this enhances the model's potential to make informed projections about future motions, even when objects are temporarily occluded. Thus, we opt for Bi-LSTMs to build our motion prediction model.

To prepare the input data for training the image and radar prediction models, we use the bounding box coordinates ($X_{min}$, $Y_{min}$, $X_{max}$, $Y_{max}$) for the camera and the point cloud centers ($X$, $Y$, $Z$) for the radar. However, to align the detections from both sensors, we deliberately use the bottom center of the bounding box ($((X_{min} + X_{max}) / 2$, $Y_{max})$ as the input to the image prediction model  (denoted as $img\_pred\_m$), while the radar point cloud centroids ($X$, $Y$) serve as the input to the radar prediction model (denoted as $rad\_pred\_m$). This intentional choice allows us to create identical network structures for the two models, thus reducing unnecessary effort.
We also create a variant of the image motion prediction model (denoted as $img\_f\_pred\_m$) that incorporates image appearance features to assess whether it improves the final prediction results. 

\begin{figure}[h!]
	\centering
	\includegraphics[width=0.45\textwidth]{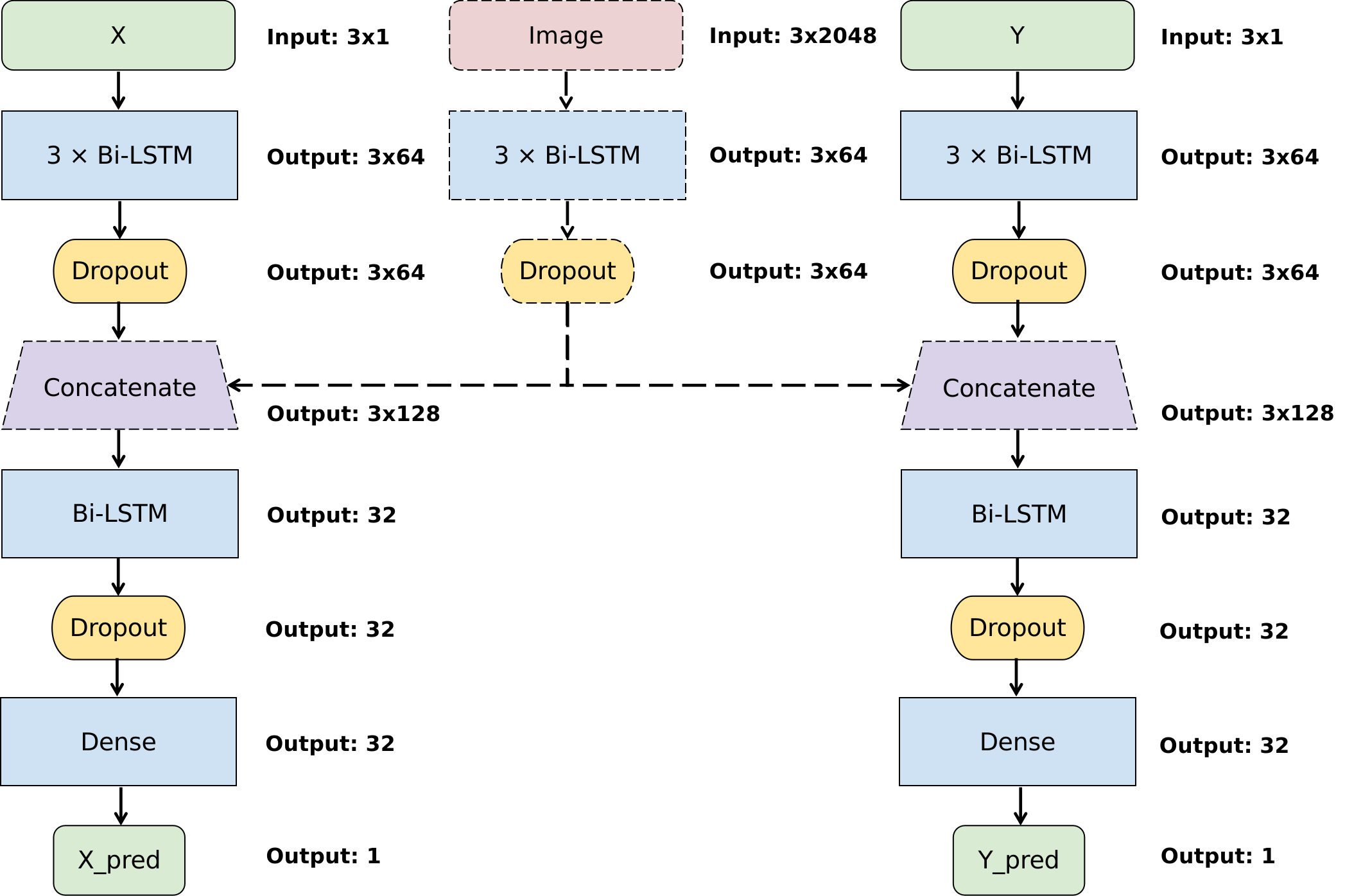}
	\caption{The network architecture of the motion model. The image data branch network is only for the image motion prediction model.}
	\label{motion_m}
\end{figure}

The network architecture of the motion model is depicted in Fig. \ref{motion_m}. 

\begin{figure}[h!]
	\centering
	\includegraphics[width=0.49\textwidth]{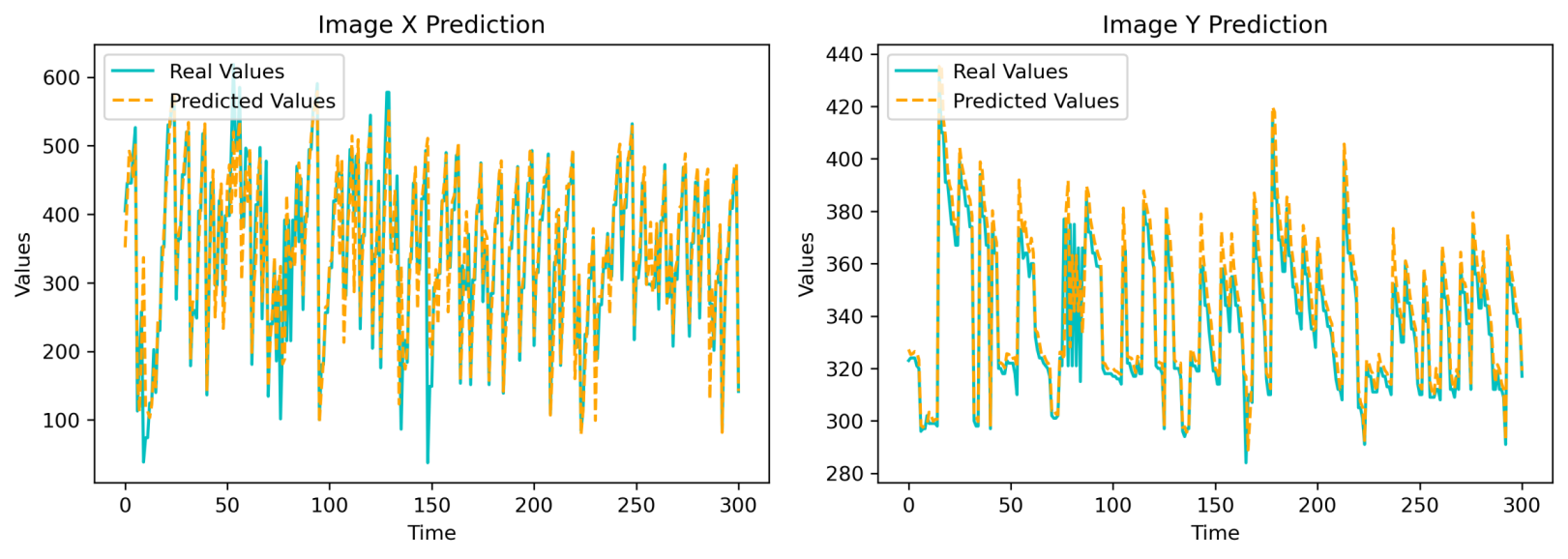}
	\caption{The prediction result of the testing dataset for the image model. Values in pixel.}
	\label{img_pm}
\end{figure}

\begin{figure}[h!]
	\centering
	\includegraphics[width=0.49\textwidth]{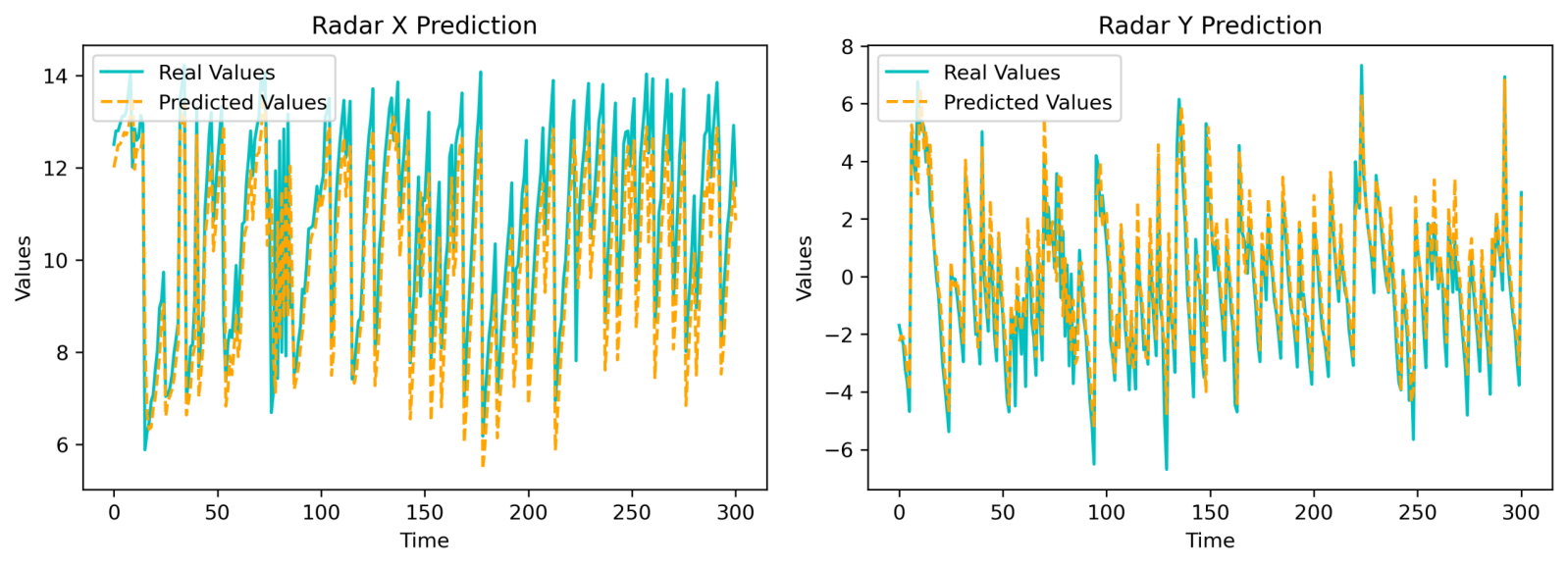}
	\caption{The prediction result of the testing dataset for the radar model. Values in meters.}
	\label{rad_pm}
\end{figure}

During training, we combine every three consecutive timesteps of the training data into a training sample, which caters to the real-time requirements of MOT and allows the prediction model to receive frequent updates to adapt to changing object movements. The predicted outputs for the testing set using the image model are displayed in Fig. \ref{img_pm}, while the corresponding radar model predictions are presented in Fig. \ref{rad_pm}. A thorough evaluation of the models' performances is provided in Table \ref{tab:my-table}, including metrics such as Root Mean Squared Error (RMSE), Mean Absolute Error (MAE), and R-squared values. Notably, the inclusion of image features does not significantly improve the performance of the image prediction model. Therefore, in our tracking algorithm, we select the model that omits image features, thereby minimizing computational overhead and accelerating model inference speed.
\begin{table}[]
\centering
\caption{The performance of the prediction model}
\label{tab:my-table}
\begin{tabular}{|l|l|l|l|l|}
\hline
\rowcolor[HTML]{C0C0C0} 
\cellcolor[HTML]{FFFFFF}                                                   & \textbf{RMSE} & \textbf{MAE} & \textbf{R-Squared} \\ \hline

\cellcolor[HTML]{EFEFEF}\textbf{img\_pred\_m \textit{(in pixels)}}   & 34.81     & 14.43    & 0.93            \\ \hline

\cellcolor[HTML]{EFEFEF}\textbf{img\_f\_pred\_m \textit{(in pixels)}} & 35.40     & 15.44    & 0.93            \\ \hline

\cellcolor[HTML]{EFEFEF}\textbf{rad\_pred\_m \textit{(in meters)}}    & 0.90      & 0.45     & 0.84            \\ \hline
\end{tabular}
\end{table}

\subsection{The Finalized MOT Method}
\begin{figure}[h!]
	\centering
	\includegraphics[width=0.49\textwidth]{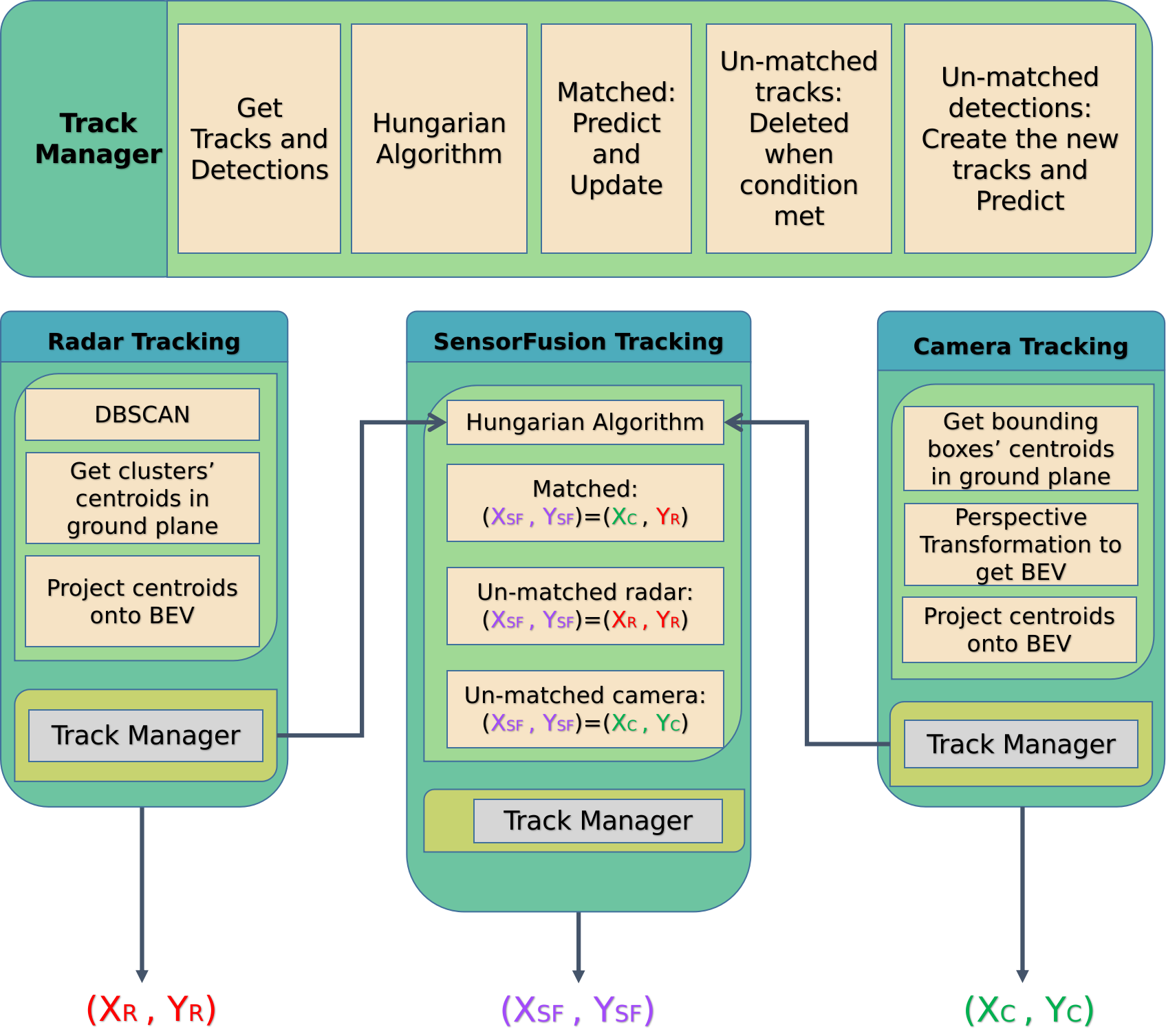}
	\caption{Algorithmic flow of the proposed MOT method with a tri-output structure: one for each sensor and one for decision-level fusion.}
	\label{pipe_realtime}
\end{figure}

%: Incorporating Appearance and Motion Models with Sensor Fusion
The finalized sensor fusion MOT method combines the appearance model to reduce ID switches and the motion prediction model for accurate trajectory forecasting. Furthermore, to ensure robustness in the face of sensor failures, this method integrates a tri-output mechanism that generates three distinct trackers: one for camera-based position ($X_C$, $Y_C$), one for radar-based position ($X_R$, $Y_R$), and one for sensor-fused position ($X_{SF}$, $Y_{SF}$). This design allows the method to continue functioning effectively, even if one sensor fails, ensuring uninterrupted tracking. The general algorithmic flow is presented in Fig. \ref{pipe_realtime}.

However, to fuse the radar and camera data for the sensor-fused tracker, we need to calibrate them first. As mentioned in Section III, the calibration involves mapping data from both the radar and the camera to a unified coordinate system. Considering the need for accurate perception of objects on the road in autonomous driving, the BEV coordinate system is been set as the unified coordinate system for integrating radar and camera data. This system anchors the radar's projection onto the ground plane as its origin and employs the ground plane as its reference plane. Similar to our previous work \cite{sengupta2022robust}, we employ an inverse perspective mapping (IPM) algorithm to transform both radar and camera data into the BEV coordinate system. By exploiting the sensor height, pitch angle, and estimated intrinsic matrix, we can achieve accurate transformations \cite{jeong2016adaptive}. 
For the objects detected using the camera, we first apply IPM to estimate a transformation matrix that generates a bird's-eye view of the scene. This transformation matrix allows us to project the bounding box centroid (i.e., the bottom center of the box) onto the ground plane, thereby obtaining the 2D spatial position of the objects detected by the camera.
For camera-detected objects, we utilize IPM to estimate a transformation matrix, projecting the bounding box centroid (bottom center) onto the ground plane, yielding 2D spatial positions.
For radar-detected objects, we employ a simple method. Instead of using a transformation matrix, we directly use cluster centroids obtained via Density-Based Spatial Clustering of Applications with Noise algorithm (DBSCAN) applied to radar point clouds, representing object locations in 2D space. This approach capitalizes on the approximate parallelism between the radar detection plane and the ground plane, reducing computational complexity. By obtaining 2D spatial positions in the BEV coordinate system from radar and camera data, we can establish associations between objects from both sensors based on their spatial proximity.

Radar excels in depth accuracy ($Y$), while cameras are superior in lateral accuracy ($X$) \cite{sengupta2022robust, ram2022fusion}. This discrepancy is due to radar's use of MIMO techniques, which have limited azimuth or lateral resolution \cite{sheeny2021radiate}. Combining both sensors through decision-level fusion compensates for their respective limitations.

Specifically, when objects are simultaneously detected and tracked by both sensors, we make $(X_{SF}, Y_{SF}) = (X_C, Y_R)$ by utilizing the depth position obtained from radar ($Y_R$) and the lateral position derived from cameras ($X_C$). In instances where objects are detected or tracked by a single sensor, we use $(X_{SF}, Y_{SF}) = (X_C, Y_C)$ or $(X_R, Y_R)$, depending on whether the object was detected or tracked by a camera or radar.

In terms of track management, each of the three trackers has its own track manager. These track managers operate independently and make their own decisions regarding when to create a new track and when to delete a lost track. Initially, in the first frame, the track manager creates a new track for each detected object in the first frame, assigning it a unique track ID. In subsequent frames, the track manager associates existing tracks with newly detected objects by solving a linear assignment problem. The linear assignment problem deals with finding an optimal assignment of a set of objects (detected objects) to another set of objects (existing tracks) with given costs associated with each assignment \cite{serratosa2015speeding}. 
To solve the linear assignment problem, we create a cost matrix $C$ to represent the association costs between detected objects and existing tracks, where $C(i, j)$ signifies the cost of associating detected object $i$ with existing track $j$.
We utilize two types of features—appearance and position—and measure their similarity distances using cosine distance for appearance ($d\_appearance(i, j)$) and Euclidean distance for position ($d\_position(i, j)$) to implement a dual-cue association strategy. We construct the cost matrix $C$ by combining these distances using a weighted sum. Specifically, we define:
\begin{equation}
\small
C(i, j) = w * d\_appearance(i, j) + (1-w) * d\_position(i, j)
\end{equation}
where $w$ is a hyperparameter that controls the relative importance of appearance and position.
Since the appearance model naturally has the advantage of distinguishing different objects, when the $d\_appearance$ is less than a lower threshold (denoted as $Thr\_low$) or greater than a higher threshold (denoted as $Thr\_high$), we directly consider two objects to be the same or different, regardless of the $d\_position$, and thus we assign the zero cost or maximum cost as the final cost. For cases other than the above, we set the weight parameter, $w$, to 0.8, striking a balance between the dominance of appearance and the influence of position. 
The Hungarian Algorithm is then used to efficiently solve the linear assignment problem and find the optimal association between detected objects and existing tracks based on the constructed cost matrix. This can result in one of three possible outcomes:
1) Matched detection-track pairs: The prediction model updates the position history list of the corresponding track with the matched detection and outputs the predicted location as the current position of the track.
2) Unmatched tracks: No associated measurements or detections exist for these tracks. The prediction model skips updating the position history list and instead predicts the track's now position, allowing for continuous estimated tracking.
3) Unmatched detections: New track IDs are assigned, and tracks are initialized. New tracks are classified as reliable if they are continuously detected for 5 frames or more, otherwise, they are classified as unreliable.

The track manager maintains three variables for each track to determine whether to delete a track:

\begin{itemize}
  \item Age ($A_t$): Incremented by 1 every frame. This variable represents the number of frames since the track was first created.
  
  \item Invisibility counter ($I_t$): Incremented by 1 every time the track is not visible in a frame. Reset to 0 when the track is visible.
  
  \item Visibility count ($V_t$): Incremented by 1 every time the object is detected by the sensor and both the predict and update stages were successfully completed.
\end{itemize}
If a track remains undetected for 20 consecutive frames (i.e., $I_t \geq 20$), the track manager removes it from consideration. Additionally, If a track's visibility score $S_t$ falls below a user-defined threshold (set at 60\% in our study), the track manager also deletes the track.
The visibility score is calculated as follows:
$$S_t = \frac{V_t}{A_t} \times 100\%$$
This visibility score provides a quantitative measure of how consistently an object has been visible and successfully tracked, serving as a decisive factor in determining the track's continued existence.

\section{Evaluate Tracking Performance}
\subsection{Tracking Data Collection and Pre-Process}
\begin{figure}[h!]
	\centering
	\includegraphics[width=0.45\textwidth]{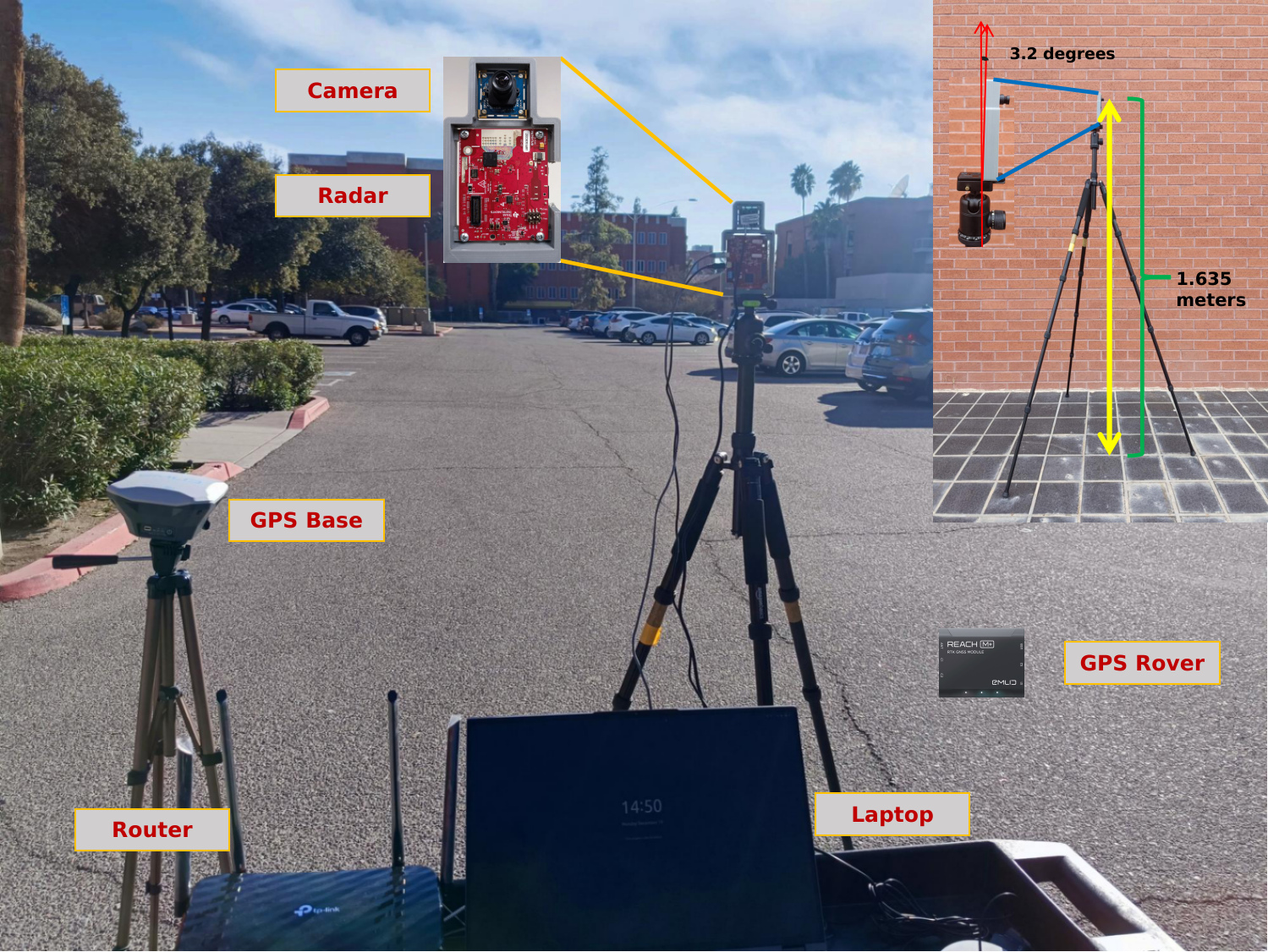}
	\caption{Data collection setup of the multi-sensor system at the parking lot.}
	\label{exp_setup}
\end{figure}

We evaluated the performance of the proposed tracking method through experiments conducted in a controlled environment at a parking lot, as depicted in Figure \ref{exp_setup}. The radar-camera system was set up on a tripod at a height of 1.635 meters, with a camera pitch angle of 3.2 degrees. The data collection was conducted under two different lighting conditions: clear visibility during the day and poor visibility at night. For each lighting condition, we have three diverse moving scenarios: (1) a single pedestrian walking forward and backward, (2) two pedestrians freely moving within the sensor's field of view, and (3) two pedestrians following crossing diagonal trajectories while a third pedestrian walks radially intersecting their paths.

Additionally, to assess the accuracy of the MOT's trajectory prediction, we employed a GPS data collection system. It comprised a base station and three rovers that communicated wirelessly. Rovers sent data to the base station, which processed it for centimeter-level precision positioning. 

The data was collected on a laptop utilizing the Robot Operating System (ROS) to configure and operate the sensors. The collected data from these sensors was then stored as ROS bag files, which contain timestamped messages from different sensors, allowing for easy synchronization and analysis of the data offline. Custom ROS packages were employed to handle specific sensors during the data collection process. For image data, the YOLOv4-Darknet image detector package was employed to generate bounding boxes for each detected object. The radar detector package was responsible for acquiring radar data and segmenting it into frames, while the DBSCAN algorithm was applied to obtain the centroids of all detected objects in each frame.
In parallel, the GPS data collector package collected WGS84 formatted GPS data from both the base station and the rovers. The data frames from the three sensors will be associated with each other by using timestamps in ROS bags. Specifically, each sensor has a timestamp array for all its data frames. A sensor with the lowest frame rate, radar in our case, its timestamp array will be traversed to find the closest matching timestamps from the other two sensors, thus associating their data frames. Given the nanosecond precision of ROS timestamps, we can empirically conclude that these sensors are well synchronized based on the described timestamp matching strategy.

\subsection{Performance Evaluation}
To evaluate our tracking approach, we used the standard CLEAR metrics \cite{bernardin2008evaluating}, including $MOTA$ (Accuracy) and $MOTP$ (Precision).
$MOTA$ measures the overall accuracy of the tracking approach by computing the ratio of correctly tracked objects to the total number of objects in the scene. It is calculated as follows:
\begin{equation}
\small
\begin{aligned}
%MOTA = 1 - \frac{\sum_{t}{FN_t} + \sum_{t}{FP_t} + \sum_{t}{IDSW_t}}{\sum_{t}{GT_t}}
MOTA &= 1 - \frac{\sum_{t=1}^{T}({FN_t} + {FP_t} + {IDSW_t})}{\sum_{t=1}^{T}{GT_t}} \\
&= 1 - \frac{\sum_{t=1}^{T}{FN_t}}{\sum_{t=1}^{T}{GT_t}}  - \frac{\sum_{t=1}^{T}{FP_t}}{\sum_{t=1}^{T}{GT_t}}  - \frac{\sum_{t=1}^{T}{IDSW_t}}{\sum_{t=1}^{T}{GT_t}} \\
\\
&= 1 - FNR  - FPR  - IDSWR
\end{aligned}
\end{equation}
where $T$ is the total number of frames, ${GT_t}$ represents the total number of objects, ${FN_t}$ is the number of missed objects (i.e., objects that were not detected), ${FP_t}$ is the number of false positives (i.e., objects that were detected but did not exist in the scene), and ${IDSW_t}$ is the number of id switches (i.e., instances where the tracking algorithm incorrectly assigned a detection to a different object) - in the $t^{th}$ frame. The ${FNR}$ (False Negative rate) is the ratio of missed detections to the total number of objects, ${FPR}$ (False Positive rate) is the ratio of non-existent object detections to the total number of objects, and ${IDSWR}$ (ID Switch rate) is the ratio of id switches to the total number of objects - across all frames.
On the other hand, $MOTP$ provides a measure of the precision of the trajectory prediction when an object is correctly detected. It evaluates the accuracy of the tracked positions compared to the ground truth positions for each object in each frame $t$. The $MOTP$ is calculated as the average distance $d_t^i$ between the tracked position and the corresponding ground truth position for the $i^{th}$ object over all frames:
\begin{equation}
%\begin{align}
MOTP = \frac{1}{N} \sum_{t=1}^{T} \sum_{i=1}^{N} d_t^i \\ 
= \frac{\sum_{t=1}^{T} \sum_{i=1}^{N} d_t^i}{\sum_{t=1}^{T}{GT_t}} 
%\end{align}     
\end{equation}
where $N$ equals the total object count in the scene, $\sum_{t=1}^{T}{GT_t}$. And $d_t^i$ is the distance between the tracked position and the ground truth position of the $i^{th}$ object in the $t^{th}$ frame. Additionally, for performance evaluation on the NuScenes dataset, we use its official AMOTA and AMOTP\cite{caesar2020nuscenes}.

\begin{figure*}[]
    \centering
    \begin{subfigure}[t]{0.49\textwidth}
        \centering
        \includegraphics[width=\textwidth]{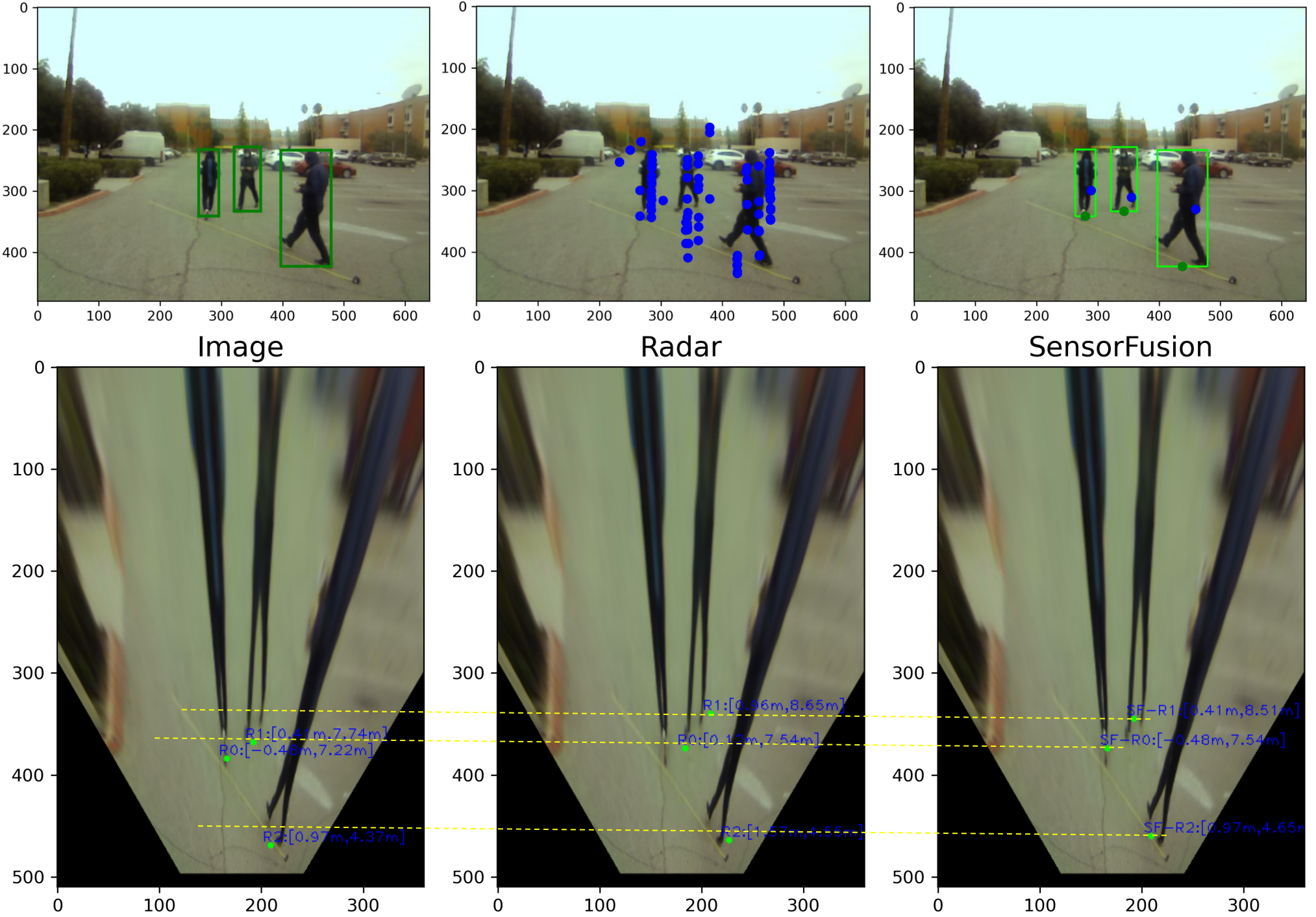}
        \caption{}
        \label{fig:image1}
    \end{subfigure}%
    \hfill
    \begin{subfigure}[t]{0.49\textwidth}
        \centering
        \includegraphics[width=\textwidth]{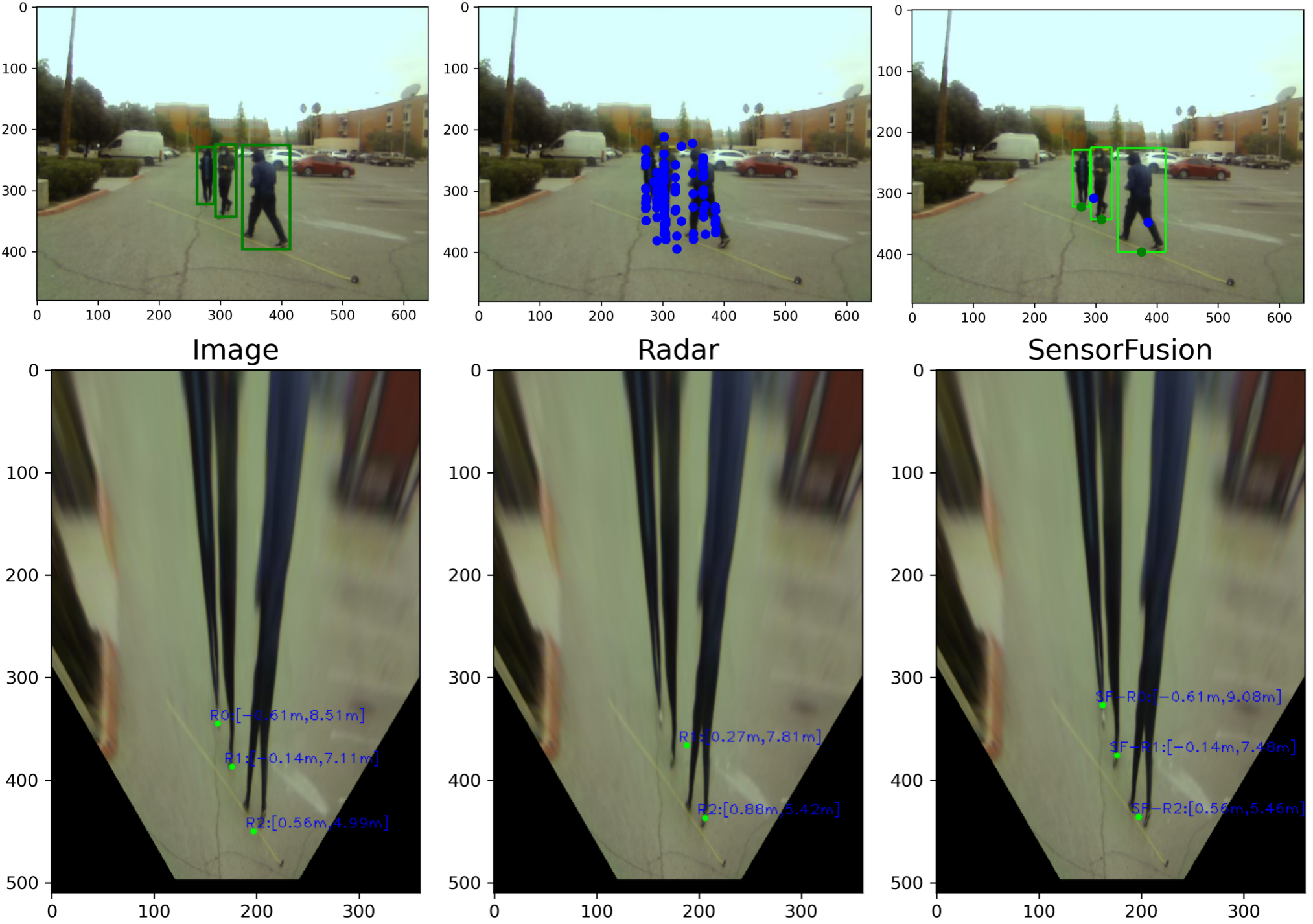}
        \caption{}
        \label{fig:image2}
    \end{subfigure}%
    
    % Second row of subfigures
    \vspace{1em} % Adjust vertical space between rows
    
    \begin{subfigure}[t]{0.49\textwidth}
        \centering
        \includegraphics[width=\textwidth]{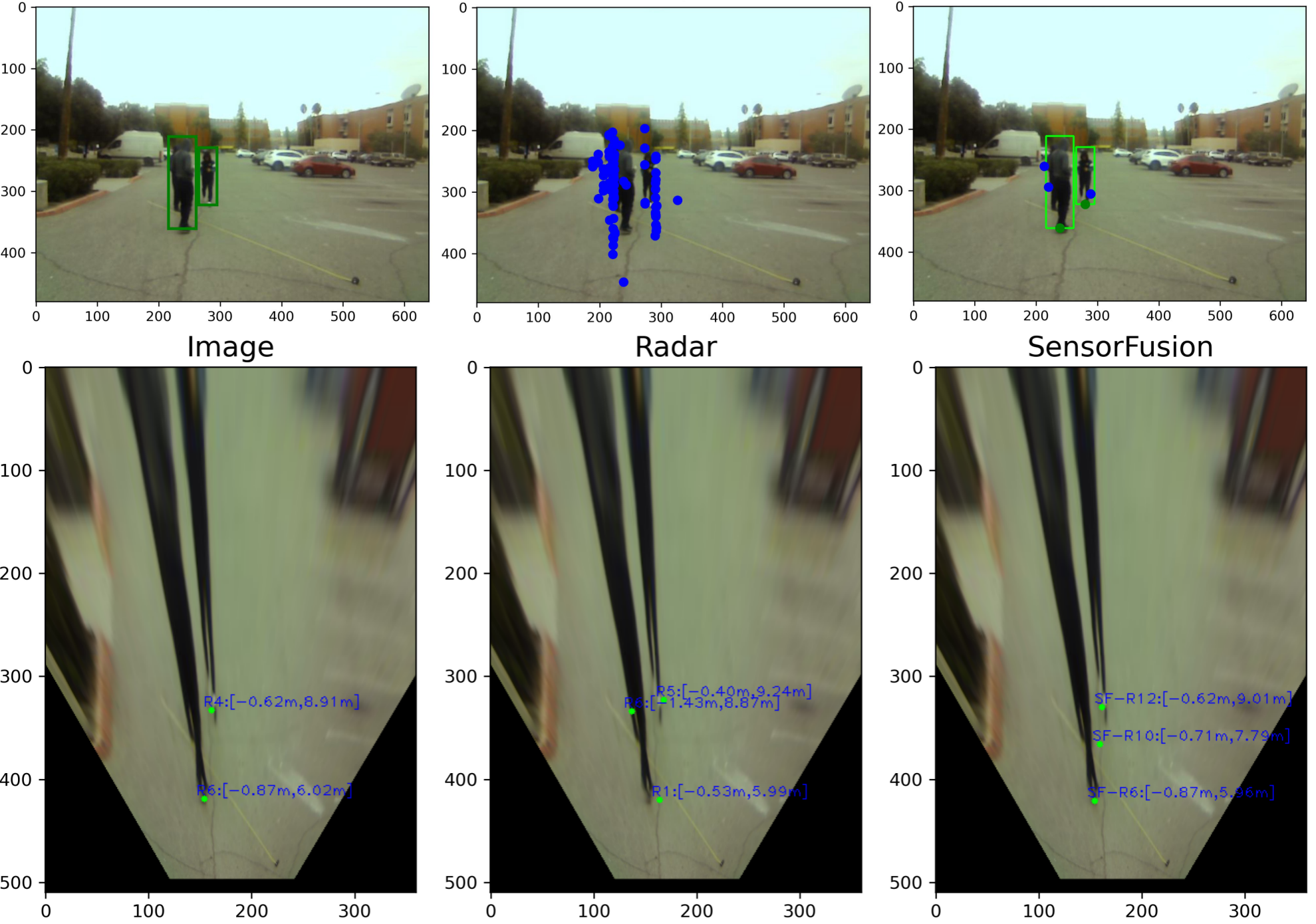}
        \caption{}
        \label{fig:image3}
    \end{subfigure}
    \hfill
    \begin{subfigure}[t]{0.49\textwidth}
        \centering
        \includegraphics[width=\textwidth]{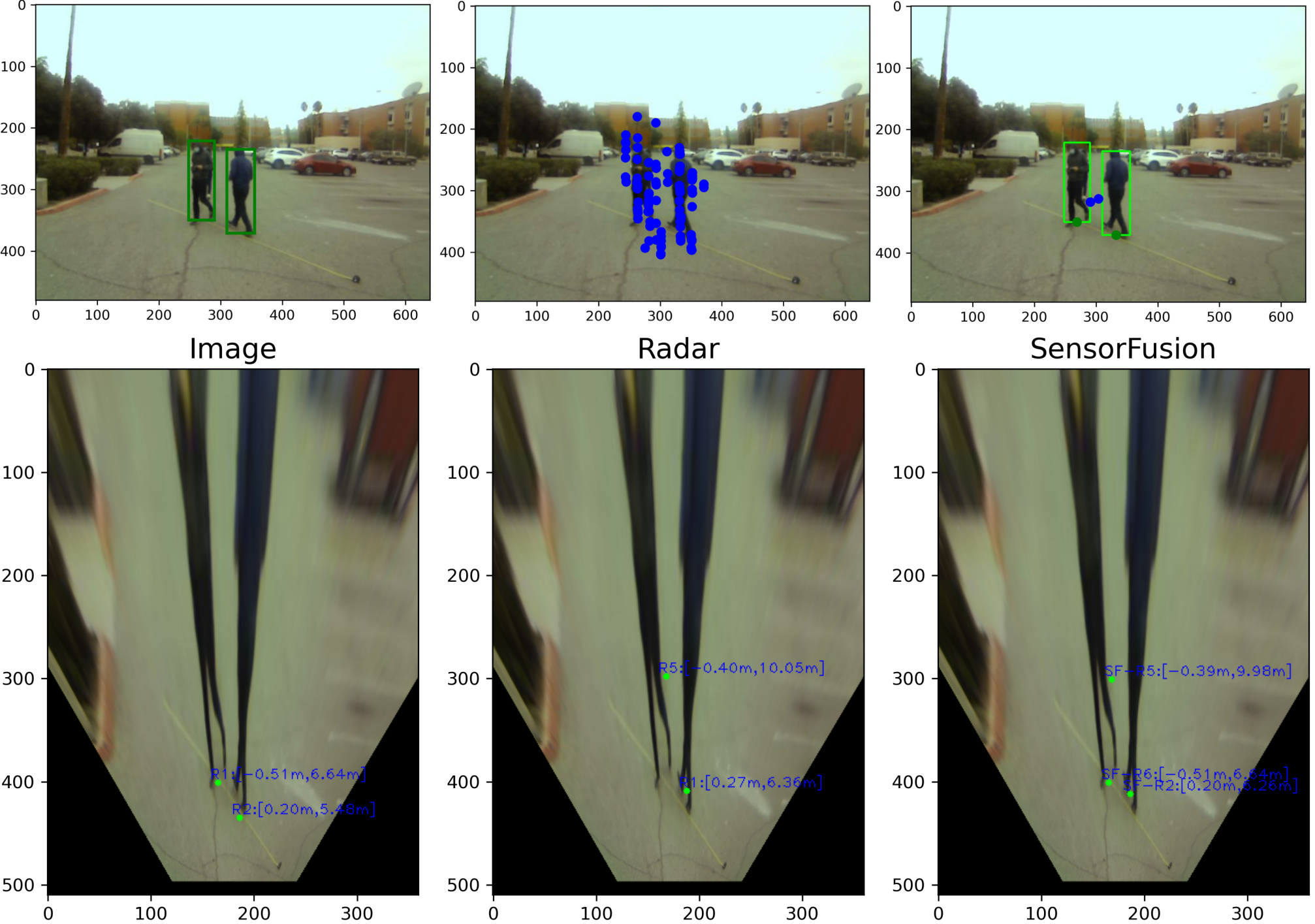}
        \caption{}
        \label{fig:image4}
    \end{subfigure}%
    %\caption{Projecting radar points onto the image using the obtained calibration matrix.}
    \label{fig:side_by_side_images}
    
    % 3rd row of subfigures
    \vspace{1em} % Adjust vertical space between rows
    \begin{subfigure}[t]{0.49\textwidth}
        \centering
        \includegraphics[width=\textwidth]{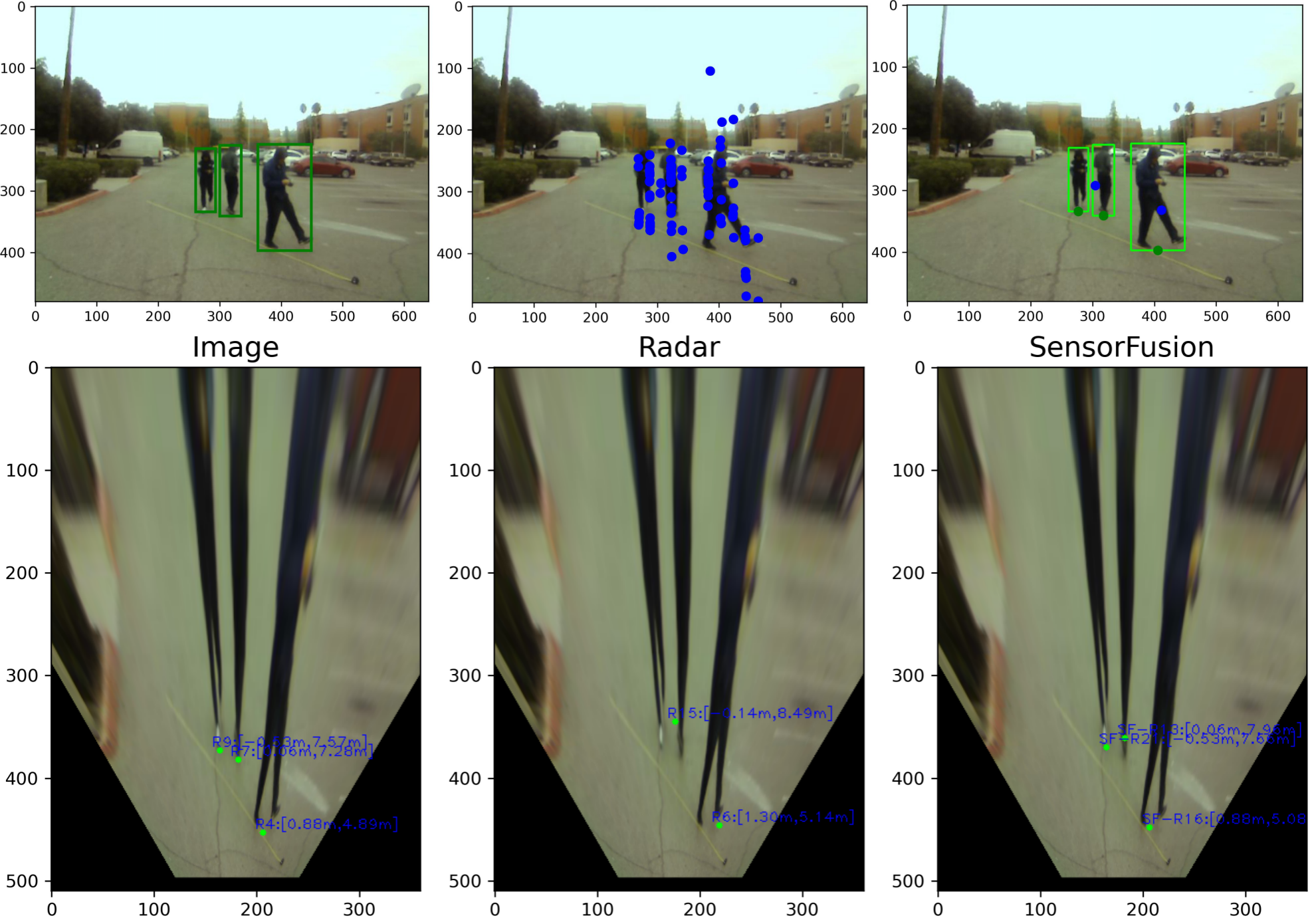}
        \caption{}
        \label{fig:image5}
    \end{subfigure}%
    \hfill
    \begin{subfigure}[t]{0.49\textwidth}
        \centering
        \includegraphics[width=\textwidth]{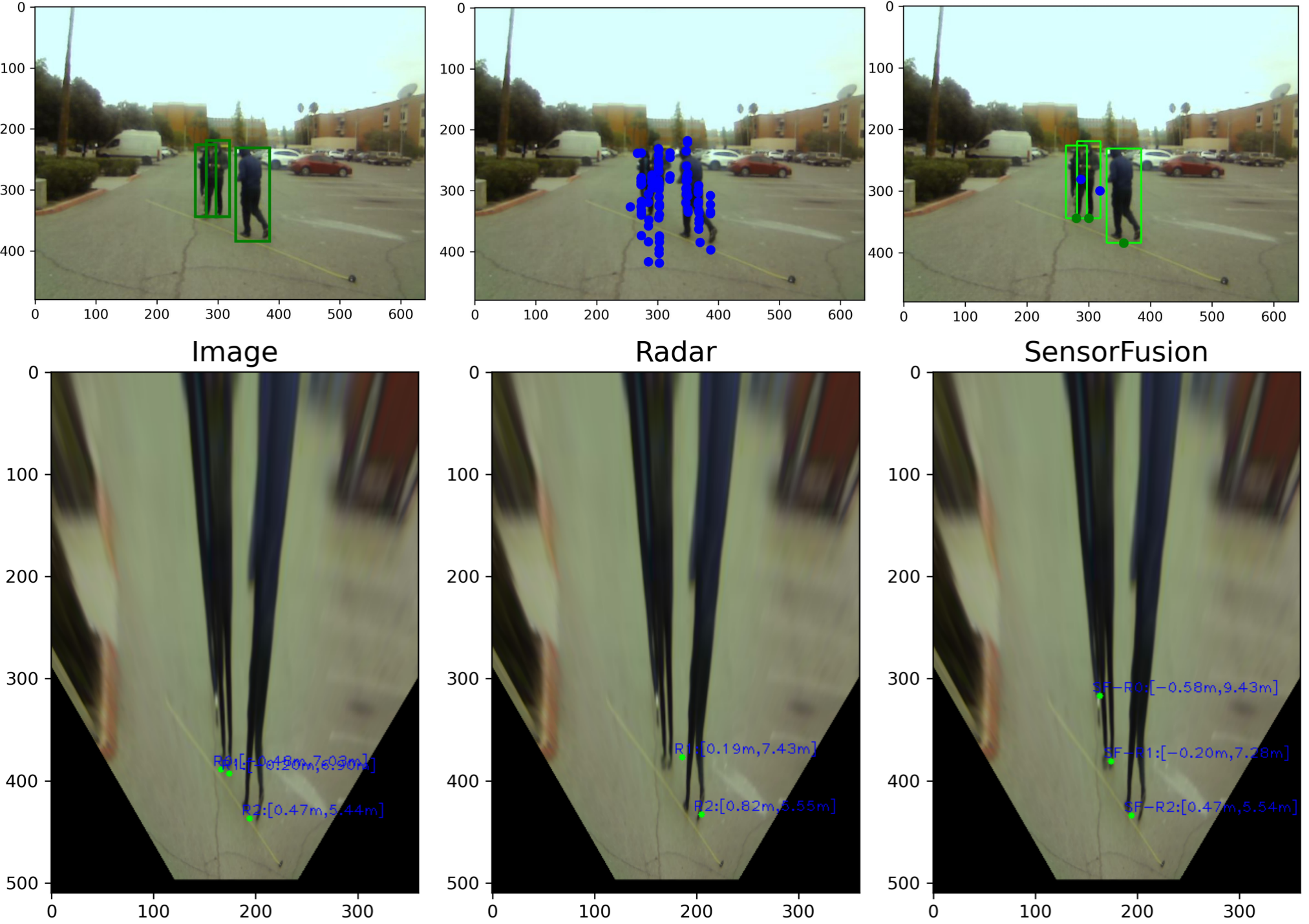}
        \caption{}
        \label{fig:image6}
    \end{subfigure}
    \caption{Multi-object Tracking examples with camera, radar, and sensor fusion. The top row corresponds to detections and their centers (or centroids) from the camera and radar, while the bottom row depicts MOT results in a bird's-eye view for camera, radar, and sensor fusion, respectively. (a) Sensor fusion MOT results when both radar and camera detect all objects in the scene. The yellow dashed lines illustrate the precision comparison between sensor fusion and the individual camera and radar MOT results.  (b) Sensor fusion MOT results when radar has missed detections while the camera detects all presented objects. (c) Sensor fusion MOT results when the camera has missed detections while radar detects all presented objects. (d) Sensor fusion MOT results when both radar and camera have missed detections. (e) Sensor fusion MOT results when radar incorrectly clusters the point clouds of closely spaced but distinct objects into a single target. (f) Sensor fusion MOT results when the camera produces non-compact bounding boxes.}
    \label{fig:all_images}
\end{figure*}

\begin{figure*}[] %htbp
	\centering
	\includegraphics[width=0.98\textwidth]{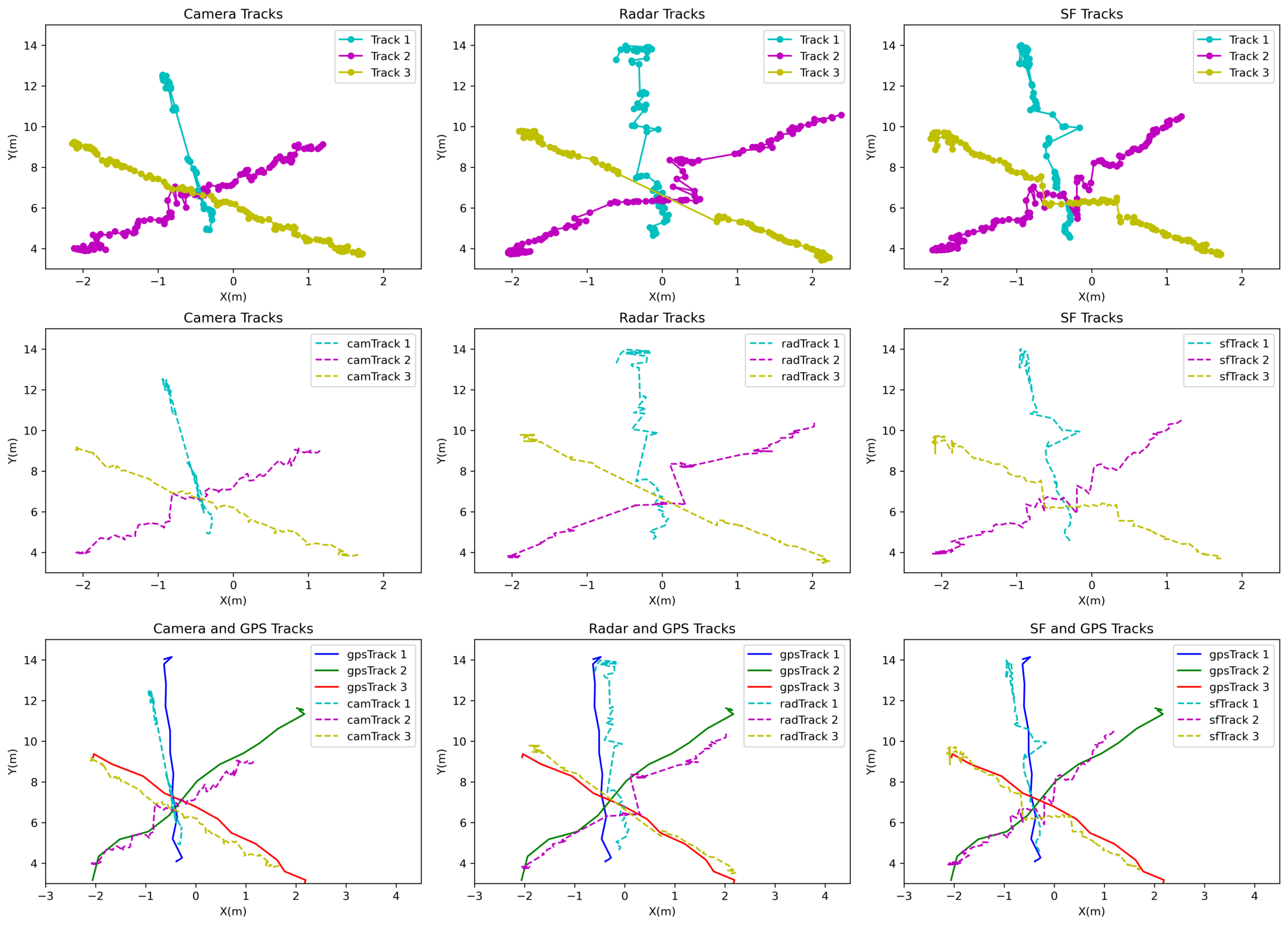}
	\caption{The trajectories corresponding to the camera, radar, and sensor fusion. The first row presents the trajectory points (indicating the positions of the objects) for three distinct objects, each obtained by the camera, radar, and sensor fusion trackers. The second row showcases the trajectories of the same three objects as tracked by the camera, radar, and sensor fusion. The third row illustrates the trajectories for these three objects generated by the camera, radar, and sensor fusion trackers, accompanied by their corresponding ground truth trajectories (obtained from GPS data).}
	\label{Tracks}
\end{figure*}

\subsection{Results and Discussion}
To evaluate the effectiveness of the proposed deep learning models, namely the appearance model and the prediction model, we conducted module-wise comparisons with traditional Kalman filter-based MOT methods, as detailed in our previous work \cite{sengupta2022robust}. Specifically, we considered four distinct configurations: The method that integrates both the appearance model and the prediction model is represented as $loc\_match+feat\_match+RNN\_pred$, while the counterpart that does not include either of these components is labeled as $loc\_match+Kalm\_pred$. The method utilizing the prediction model but not the appearance model is referred to as $loc\_match+RNN\_pred$, while the alternative employing the appearance model but not the prediction model is designated as $loc\_match+feat\_match+Kalm\_pred$.
The comparisons are presented in Table. \ref{tab4}, using the CLEAR metrics under Scenario 3. From the table, it is evident that the sensor fusion tracker outperforms individual sensors concerning MOTP accuracy. However, in terms of MOTA accuracy, the sensor fusion tracker does not exhibit a significant advantage over the individual sensor trackers. The slightly lower MOTA accuracy of the sensor fusion tracker can be blamed for its excessively high FPR (incorrectly showing trajectories that do not correspond to any objects). This elevated FPR in the sensor fusion tracker can be attributed to two main factors: 1) the sensor fusion tracker incorporates all incorrectly detected objects from both sensors, and 2) the objects detected by the camera and radar are not effectively associated. A more sophisticated matching strategy or track management approach can help mitigate this problem. Despite this limitation, the sensor fusion tracker presents itself as a promising MOT solution relative to the single-sensor tracker. 

Our proposed method (i.e., $loc\_match+feat\_match+RNN\_pred$), which combines location matching, feature matching, and RNN(Bi-LSTM) prediction, achieves the highest MOTA accuracy, particularly when compared to the Kalman filter-based approaches (i.e., $loc\_match+Kalm\_pred$). The appearance model plays a vital role in our method, as it can better match objects, leading to direct improvements in MOTA (with noticeable improvements in FNR, FPR, and IDSWR) and indirect improvements in MOTP accuracy. However, the RNN prediction model does not perform better than the Kalman filter as expected. Several factors contribute to this outcome. Firstly, we trained the prediction model using only three timesteps of data, limiting its ability to capture long-term motion information for objects. Secondly, the prediction model lacks adaptability. The Kalman filter continuously updates its state matrix using new detections, while the RNNs are trained offline, once deployed, they operate with fixed weights. This static nature may hinder their ability to capture the dynamically changing motions inherent in the tracking task. Despite its limitations, the proposed motion prediction model highlights the promise of deep learning. Deep learning models, with their ability to learn from data directly, can evolve further with larger datasets and improved computational power. Unlike traditional methods dependent on human-crafted features, deep learning excels at generalizing to unseen scenarios, making it highly valuable for tasks like motion prediction.

Furthermore, a series of tracking results is provided in Fig. \ref{fig:all_images}, demonstrating the robustness and precision of the proposed sensor fusion-based MOT method. In Fig. \ref{fig:image1}, it is evident that the sensor fusion tracker achieves more accurate object localization compared to the camera and radar trackers. Fig. \ref{fig:image2}, Fig. \ref{fig:image3}, and Fig. \ref{fig:image4} showcase the sensor fusion tracker's ability to accurately locate objects even when either the camera, radar, or both are not functioning correctly (i.e., failing to provide object detections).
Fig. \ref{fig:image5} corresponds to a scenario where the radar can detect all objects in the scene but mistakenly clusters them into a single object due to their close proximity, a common occurrence in radar detection and a primary reason for high radar FNR. However, the sensor fusion tracker effectively handles this situation.
Fig. \ref{fig:image6} represents a scenario where the camera can detect all appearing objects, but the generated bounding boxes are not sufficiently compact, resulting in lower localization accuracy. This is a significant contributing factor to the reduced precision in the camera tracker. Nevertheless, the sensor fusion tracker adeptly addresses this challenge.

Fig. \ref{Tracks} presents trajectories of different objects corresponding to the camera, radar, and sensor fusion to further emphasize the superiority of the proposed method. From the three plots in the first row, we can observe that both the camera and radar trajectories exhibit non-realistic tracks with missing detections. Some segments of these tracks (represented as long straight lines on the trajectory plot) do not intersect with any points. This phenomenon is particularly noticeable in Track1 in the Camera Tracks plot and Track3 in the Radar Tracks plot. This occurs because the camera generates images that describe a two-dimensional space and inherently struggles to distinguish overlapping or occluded objects. In this experiment, individuals walking radially are frequently occluded by those crossing their paths, resulting in numerous missing detections and the creation of the unrealistic Track1. Similarly, for radar, its angular resolution is generally lower, and when two targets are very close in space, their corresponding point clouds are prone to being clustered as a single object, leading to missing detections (e.g., Track1 and Track3 in Radar Tracks plot) or erroneous tracking (e.g., Track2 in Radar Tracks plot). However, sensor fusion effectively addresses this issue by integrating detections from both the camera and radar, as visually demonstrated in the second row of three images. Also, for the three plots in the third row, through a comparison of the trajectories generated by the camera, radar, and sensor fusion with the ground truth trajectories, we can observe that the camera can provide relatively accurate lateral positioning but lower depth positioning accuracy (the trajectories are shorter than the ground truth trajectories). In contrast, radar can provide more accurate depth positioning but lower lateral positioning accuracy (the trajectories deviate laterally from the ground truth trajectories). Sensor fusion, on the other hand, achieves improved accuracy in both lateral and depth positions by fusing the data from the camera and radar.

\begin{table*}[]
\centering
\caption{CLEAR-MOT Metrics for Modularization Performance Comparison}
\label{tab4}
\renewcommand{\arraystretch}{1.5}
\resizebox{\textwidth}{!}{%
\begin{tabular}{|l|c|c|c|c|c|c|c|c|c|c|c|c|c|c|c|}
\hline
 & \multicolumn{3}{c|}{\cellcolor[HTML]{C0C0C0}FPR(\%)} & \multicolumn{3}{c|}{\cellcolor[HTML]{C0C0C0}FNR(\%)} & \multicolumn{3}{c|}{\cellcolor[HTML]{C0C0C0}IDSWR(\%)} & \multicolumn{3}{c|}{\cellcolor[HTML]{C0C0C0}MOTA(\%)} & \multicolumn{3}{c|}{\cellcolor[HTML]{C0C0C0}MOTP(m)} \\ \cline{2-16} 
\multirow{-2}{*}{} & \multicolumn{1}{l|}{\cellcolor[HTML]{EFEFEF}Cam} & \multicolumn{1}{l|}{\cellcolor[HTML]{EFEFEF}Rad} & \multicolumn{1}{l|}{\cellcolor[HTML]{EFEFEF}SF} & \multicolumn{1}{l|}{\cellcolor[HTML]{EFEFEF}Cam} & \multicolumn{1}{l|}{\cellcolor[HTML]{EFEFEF}Rad} & \multicolumn{1}{l|}{\cellcolor[HTML]{EFEFEF}SF} & \multicolumn{1}{l|}{\cellcolor[HTML]{EFEFEF}Cam} & \multicolumn{1}{l|}{\cellcolor[HTML]{EFEFEF}Rad} & \multicolumn{1}{l|}{\cellcolor[HTML]{EFEFEF}SF} & \multicolumn{1}{l|}{\cellcolor[HTML]{EFEFEF}Cam} & \multicolumn{1}{l|}{\cellcolor[HTML]{EFEFEF}Rad} & \multicolumn{1}{l|}{\cellcolor[HTML]{EFEFEF}SF} & \multicolumn{1}{l|}{\cellcolor[HTML]{EFEFEF}Cam} & \multicolumn{1}{l|}{\cellcolor[HTML]{EFEFEF}Rad} & \multicolumn{1}{l|}{\cellcolor[HTML]{EFEFEF}SF} \\ \hline
\cellcolor[HTML]{EFEFEF}loc\_match+Kalm\_pred method & 15.13 & 6.44 & 16.46 & 7.77 & 8.99 & 1.84 & 1.33 & 0.92 & 1.84 & 75.77 & 83.65 & 79.86 & 0.43 & 0.48 & 0.33 \\ \hline
\cellcolor[HTML]{EFEFEF}loc\_match+feat\_match+Kalm\_pred method & 7.29 & 6.44 & 9.22 & 4.52 & 8.99 & 1.21 & 0.49 & 0.92 & 0.84 & 87.70 & 83.65 & 88.73 & 0.41 & 0.48 & 0.28 \\ \hline
\cellcolor[HTML]{EFEFEF}loc\_match+RNN\_pred method & 10.23 & 5.92 & 11.44 & 6.40 & 9.05 & 0.67 & 0.95 & 0.88 & 1.42 & 82.42 & 84.15 & 86.47 & 0.45 & 0.50 & 0.36 \\ \hline
\cellcolor[HTML]{EFEFEF}loc\_match+feat\_match+RNN\_pred method & 4.20 & 5.92 & 9.77 & 5.06 & 9.05 & 0.55 & 0.26 & 0.88 & 0.66 & 90.48 & 84.15 & 89.02 & 0.44 & 0.50 & 0.34 \\ \hline
\end{tabular}%
}
\end{table*}

%%%%%%%%%%%%%%%%%%%%%%%%%%%%%
\begin{table*}[]
\centering
\caption{Performance of the Proposed Method Across Five Different Tracking Scenarios}
\label{tab5}
\renewcommand{\arraystretch}{1.5}
\resizebox{\textwidth}{!}{%
\begin{tabular}{|l|c|c|c|c|c|c|c|c|c|c|c|c|c|c|c|}
\hline
 & \multicolumn{3}{c|}{\cellcolor[HTML]{C0C0C0}FPR(\%)} & \multicolumn{3}{c|}{\cellcolor[HTML]{C0C0C0}FNR(\%)} & \multicolumn{3}{c|}{\cellcolor[HTML]{C0C0C0}IDSWR(\%)} & \multicolumn{3}{c|}{\cellcolor[HTML]{C0C0C0}MOTA(\%)} & \multicolumn{3}{c|}{\cellcolor[HTML]{C0C0C0}MOTP(m)} \\ \cline{2-16} 
\multirow{-2}{*}{} & \multicolumn{1}{l|}{\cellcolor[HTML]{EFEFEF}Cam} & \multicolumn{1}{l|}{\cellcolor[HTML]{EFEFEF}Rad} & \multicolumn{1}{l|}{\cellcolor[HTML]{EFEFEF}SF} & \multicolumn{1}{l|}{\cellcolor[HTML]{EFEFEF}Cam} & \multicolumn{1}{l|}{\cellcolor[HTML]{EFEFEF}Rad} & \multicolumn{1}{l|}{\cellcolor[HTML]{EFEFEF}SF} & \multicolumn{1}{l|}{\cellcolor[HTML]{EFEFEF}Cam} & \multicolumn{1}{l|}{\cellcolor[HTML]{EFEFEF}Rad} & \multicolumn{1}{l|}{\cellcolor[HTML]{EFEFEF}SF} & \multicolumn{1}{l|}{\cellcolor[HTML]{EFEFEF}Cam} & \multicolumn{1}{l|}{\cellcolor[HTML]{EFEFEF}Rad} & \multicolumn{1}{l|}{\cellcolor[HTML]{EFEFEF}SF} & \multicolumn{1}{l|}{\cellcolor[HTML]{EFEFEF}Cam} & \multicolumn{1}{l|}{\cellcolor[HTML]{EFEFEF}Rad} & \multicolumn{1}{l|}{\cellcolor[HTML]{EFEFEF}SF} \\ \hline
\cellcolor[HTML]{EFEFEF}Scenario 1 (Daytime and Nighttime) & - & - & - & - & - & - & - & - & - & - & - & - & 0.34 & 0.22 & 0.31 \\ \hline
\cellcolor[HTML]{EFEFEF}Scenario 2 (Daytime) & 3.31 & 5.54 & 6.87 & 3.24 & 4.65 & 0.65 & 0.20 & 0.42 & 0.73 & 93.25 & 89.39 & 91.75 & 0.39 & 0.52 & 0.34 \\ \hline
\cellcolor[HTML]{EFEFEF}Scenario 2 (Nighttime) & 5.45 & 6.64 & 11.52 & 7.62 & 5.05 & 0.37 & 1.23 & 0.99 & 1.80 & 85.70 & 87.32 & 86.31 & 0.45 & 0.43 & 0.38 \\ \hline
\cellcolor[HTML]{EFEFEF}Scenario 3 (Daytime) & 4.20 & 5.92 & 9.77 & 5.06 & 9.05 & 0.55 & 0.26 & 0.88 & 0.66 & 90.48 & 84.15 & 89.02 & 0.44 & 0.50 & 0.34 \\ \hline
\cellcolor[HTML]{EFEFEF}Scenario 3 (Nighttime) & 7.17 & 8.82 & 12.33 & 9.95 & 7.61 & 0.90 & 2.02 & 1.14 & 2.56 & 80.86 & 82.43 & 84.21 & 0.48 & 0.51 & 0.39 \\ \hline
\cellcolor[HTML]{EFEFEF}\textbf{\centering \ \ \ \ \ \ \ \ \ \ \ \ \ \ \ \ Average} & \textbf{5.03} & \textbf{6.48} & \textbf{10.12} & \textbf{6.46} & \textbf{6.59} & \textbf{0.61} & \textbf{0.92} & \textbf{0.85} & \textbf{1.43} & \textbf{87.57} & \textbf{85.82} & \textbf{87.82} & \textbf{0.42} & \textbf{0.43} & \textbf{0.35} \\ \hline
\end{tabular}%
}
\end{table*}
%%%%%%%%%%%%%#####################
\begin{table*}[]
\centering
\caption{Performance of the Method \cite{cui2023online} Across Five Different Tracking Scenarios}
\label{tab:online_method}
\renewcommand{\arraystretch}{1.5}
\resizebox{\textwidth}{!}{%
\begin{tabular}{|l|c|c|c|c|c|c|c|c|c|c|c|c|c|c|c|}
\hline
 & \multicolumn{3}{c|}{\cellcolor[HTML]{C0C0C0}FPR(\%)} & \multicolumn{3}{c|}{\cellcolor[HTML]{C0C0C0}FNR(\%)} & \multicolumn{3}{c|}{\cellcolor[HTML]{C0C0C0}IDSWR(\%)} & \multicolumn{3}{c|}{\cellcolor[HTML]{C0C0C0}MOTA(\%)} & \multicolumn{3}{c|}{\cellcolor[HTML]{C0C0C0}MOTP(m)} \\ \cline{2-16} 
\multirow{-2}{*}{} & \cellcolor[HTML]{EFEFEF}Cam & \cellcolor[HTML]{EFEFEF}Rad & \cellcolor[HTML]{EFEFEF}SF & \cellcolor[HTML]{EFEFEF}Cam & \cellcolor[HTML]{EFEFEF}Rad & \cellcolor[HTML]{EFEFEF}SF & \cellcolor[HTML]{EFEFEF}Cam & \cellcolor[HTML]{EFEFEF}Rad & \cellcolor[HTML]{EFEFEF}SF & \cellcolor[HTML]{EFEFEF}Cam & \cellcolor[HTML]{EFEFEF}Rad & \cellcolor[HTML]{EFEFEF}SF & \cellcolor[HTML]{EFEFEF}Cam & \cellcolor[HTML]{EFEFEF}Rad & \cellcolor[HTML]{EFEFEF}SF \\ \hline
\cellcolor[HTML]{EFEFEF}Scenario 1 (Daytime and Nighttime) & - & - & - & - & - & - & - & - & - & - & - & - & 0.34 & 0.22 & 0.31 \\ \hline
\cellcolor[HTML]{EFEFEF}Scenario 2 (Daytime) & 3.48 & 5.82 & 7.21 & 3.26 & 4.75 & 0.66 & 0.24 & 0.50 & 0.88 & 93.02 & 88.93 & 91.25 & 0.41 & 0.53 & 0.34 \\ \hline
\cellcolor[HTML]{EFEFEF}Scenario 2 (Nighttime) & 5.72 & 6.97 & 12.10 & 7.66 & 5.16 & 0.37 & 1.48 & 1.19 & 2.16 & 85.14 & 86.68 & 85.37 & 0.46 & 0.45 & 0.40 \\ \hline
\cellcolor[HTML]{EFEFEF}Scenario 3 (Daytime) & 4.41 & 6.22 & 10.26 & 5.08 & 9.24 & 0.56 & 0.31 & 1.06 & 0.79 & 90.20 & 83.48 & 88.39 & 0.46 & 0.50 & 0.34 \\ \hline
\cellcolor[HTML]{EFEFEF}Scenario 3 (Nighttime) & 7.53 & 9.26 & 12.95 & 10.00 & 7.77 & 0.91 & 2.42 & 1.37 & 3.07 & 80.05 & 81.60 & 83.07 & 0.50 & 0.51 & 0.41 \\ \hline
\cellcolor[HTML]{EFEFEF}\textbf{\centering \ \ \ \ \ \ \ \ \ \ \ \ \ \ \ \ Average} & \textbf{5.28} & \textbf{7.07} & \textbf{10.63} & \textbf{6.50} & \textbf{6.73} & \textbf{0.62} & \textbf{1.11} & \textbf{1.03} & \textbf{1.73} & \textbf{87.10} & \textbf{85.17} & \textbf{87.02} & \textbf{0.43} & \textbf{0.44} & \textbf{0.36} \\ \hline
\end{tabular}%
}
\end{table*}
%%%%%%%%%%%%%#####################
\begin{table*}[]
\centering
\caption{Sensor Fusion Tracking Performance Comparison on NuScenes Test Set for Car and Pedestrian}
\label{tab:nus_method}
\renewcommand{\arraystretch}{1.5}
\resizebox{0.7\textwidth}{!}{%
\begin{tabular}{|l|c|c|c|c|c|c|c|c|c|c|c|c|c|c|c|}
\hline
 & {\cellcolor[HTML]{C0C0C0}FP} & {\cellcolor[HTML]{C0C0C0}FN} & {\cellcolor[HTML]{C0C0C0}IDSW} &{\cellcolor[HTML]{C0C0C0}AMOTA} & {\cellcolor[HTML]{C0C0C0}AMOTP}  & {\cellcolor[HTML]{C0C0C0}FPS} \\ \cline{2-7} 
 \hline
\cellcolor[HTML]{EFEFEF}Our Method+\cite{kim2023crn} & 11051 & 11205 & 975 & 54.5 & 50.7 & 8.2 \\ \hline
\cellcolor[HTML]{EFEFEF}Combined Method(\cite{kim2023crn}+\cite{li2023poly}) & 10582 & 12313 & 1022 & 51.2 & 56.2 & 9.1 \\ \hline
\end{tabular}%
}
\end{table*}

We also evaluated our proposed MOT method on different scenarios, and the performance is summarized in Table \ref{tab5}. According to the results, Scenario 1 (with only one person) has the best MOTP, possibly due to the fact that it merely involves radial motion, resulting in minimal lateral errors. Notably, MOTA wasn't computed for Scenario 1 due to its single-object tracking nature. In contrast, Scenario 2 (with two people) has slightly higher MOTA than Scenario 3, likely because Scenario 2 only involves two people, and they have fewer opportunities to block each other's movement, resulting in fewer occasions for the tracker to lose track of them. In contrast, Scenario 3 involves three people, and their walking paths are deliberately arranged to intersect with each other multiple times, increasing the likelihood of occlusions and making it more difficult for the tracker to maintain accurate tracking. Also, the individual sensor trackers in Scenario 2 displayed slightly better MOTP accuracy than those in Scenario 3, reflecting the reduced complexity of tracking two pedestrians. However, the sensor fusion tracker maintained consistent MOTP accuracy in both scenarios, highlighting its capability to integrate data from multiple sensors, thus mitigating track discontinuities and enhancing MOTP precision. Lastly, the nighttime scenario indeed leads to some redundant detections and missed detections by the camera, causing a decrease in tracking performance. Nevertheless, the decline is less pronounced than anticipated, potentially owing to the fact that our testing environment was not completely dark, still with some lighting present in the parking lot. 

To further evaluate our approach, Method \cite{cui2023online} was benchmarked against our method on our dataset. The full performance metrics of Method \cite{cui2023online}, as presented in Table \ref{tab:online_method}, when compared with those in Table \ref{tab5}, reveal a marginal superiority of our methodology. We argue that the efficacy of our method stems from the employment of an appearance model that more accurately gauges object similarity through similarity scoring rather than treating the similarity assessment as a classification task, as seen in Method \cite{cui2023online}. Furthermore, our dual-cue association strategy based on the Hungarian algorithm, which integrates appearance similarity scores with positional distance, achieves a more global optimal matching of detections to tracks compared to Method \cite{cui2023online}'s sequential graph-based association followed by Hungarian matching. Our calibration technique and the bird’s-eye view transformation, based on IPM, also bolster MOTP.
\begin{figure*}[h!]
	\centering
	\includegraphics[width=0.98\textwidth]{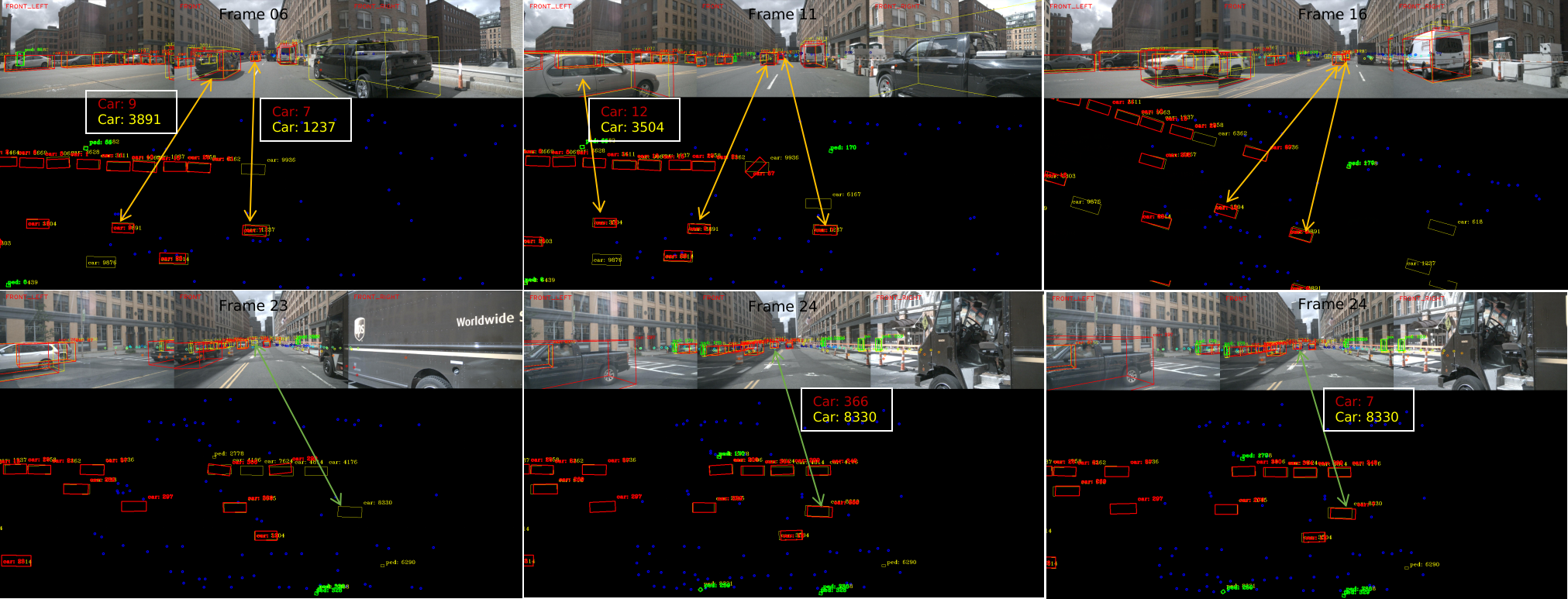}
	\caption{Sensor fusion tracking results on the NuScenes dataset using our proposed method. Each frame includes both camera and bird's-eye view visualization, with radar points projected onto them as colored circles. Ground truth is marked in yellow, and our method's results are in red. The second Frame 24 shows the combined method's tracking results.}
	\label{nus_track}
\end{figure*}
\begin{figure*}[h!]
	\centering
	\includegraphics[width=0.98\textwidth]{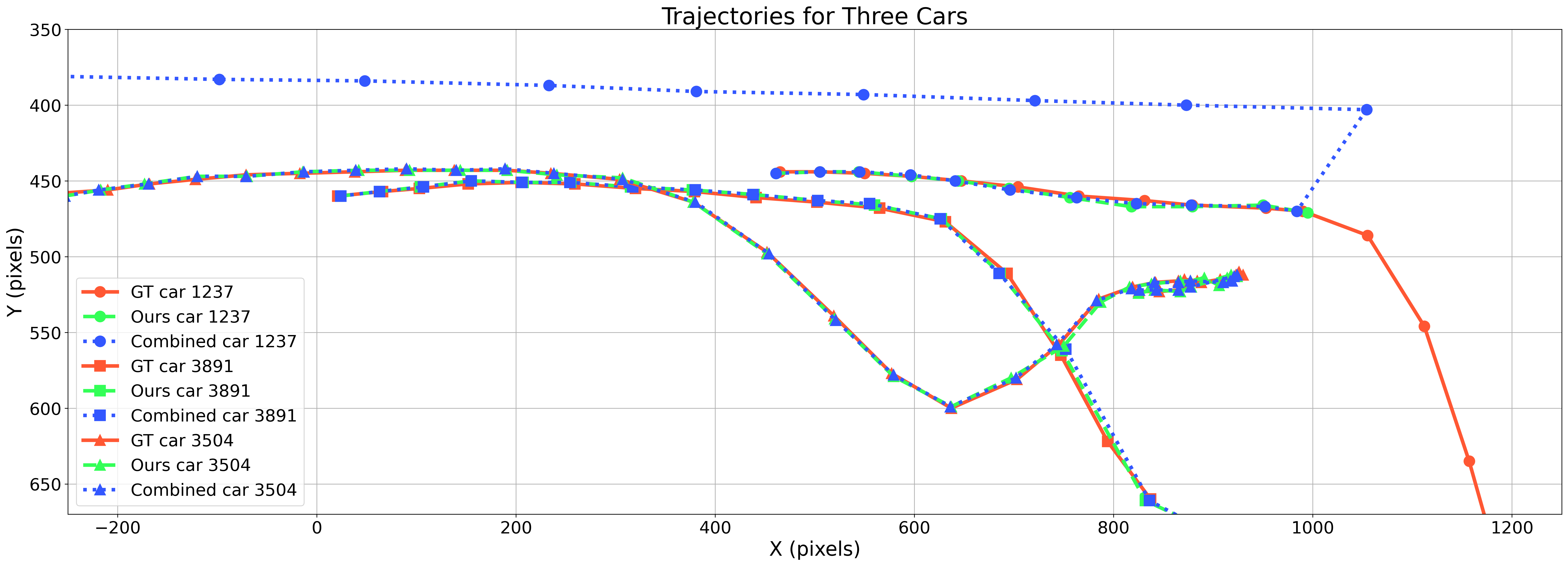}
	\caption{Trajectories of three cars (with ground truth tracking IDs: 1237, 3891, and 3504) from the Fig. \ref{nus_track}, generated using our proposed method and the combined method. The ground truth trajectories are derived from the NuScenes dataset.}
	\label{nus_tracjectory}
\end{figure*}

Finally, we assessed our method’s performance on the public dataset NuScenes. Here, we contrast our method with a joint approach comprising Methods \cite{kim2023crn} and \cite{li2023poly}: the former processes data to yield radar-camera fusion detection results, while the latter contributes to the tracking process. For a fair comparison, our method uses Method \cite{kim2023crn} as the detector and is adapted from Method \cite{li2023poly}'s framework, integrating our appearance model, substituting the original EKF with our Bi-LSTM motion model, and adopting a dual-cue association strategy centered on the Hungarian algorithm over the original two-stage data association strategy. It is pertinent to note that our evaluation was confined to the 'car' and 'pedestrian' categories, reflecting our appearance model's training, hence multi-category optimizations of Method \cite{li2023poly} may not be fully realized in this assessment. From Table \ref{tab:nus_method}, our method shows reduced ID switches and an enhanced AMOTA owing to the integration of the appearance model. Impressively, our method also demonstrates superior AMOTP, affirming the Bi-LSTM motion model's advantage over EKF in estimating motion. This efficacy is likely attributed to the Bi-LSTM motion model's extensive training on substantial datasets, equipping it to adeptly handle the irregular and stochastic motions prevalent in real-world scenarios.
Moreover, our method operates at 14 frames per second (FPS) on our dataset and, following integration with Method \cite{kim2023crn}, functions at 8 FPS on the NuScenes dataset. This performance is achieved on a system with an Xeon E5-2695v4 4-core CPU and an NVIDIA P100 GPU. It has the potential to reach a real-time frame rate with 1) optimized algorithmic implementation; and 2) higher processing power to enable faster detector throughput.
Fig. \ref{nus_track} illustrates the sensor fusion tracking results of our proposed method. It is evident from the figure that the bounding boxes generated by our method are generally well-aligned with the ground truth. Additionally, our tracking results maintain consistent tracking IDs across multiple frames, demonstrating the accuracy and reliability of our method.
Fig. \ref{nus_tracjectory} further presents the trajectories of the three cars (with ground truth tracking IDs: 1237, 3891, and 3504) depicted in Fig. \ref{nus_track}, generated by our proposed method and the combined method. The ground truth trajectories are derived from the NuScenes dataset. As shown in the figure, the trajectories produced by our method are closely aligned with the ground truth trajectories. For car 3891, our method provides a better fit to the ground truth trajectory compared to the combined method. And, for car 3504, the combined method exhibits a clear mismatch, while our method, despite missing the object, avoids incorrect associations.
The reason for the misassociation in the combined method can be seen in the bottom three images of Fig. \ref{nus_track}. In Frame 23, a new detection (Car: 8330) moving in the opposite direction appears near an existing trajectory (Car: 7). As described in Section IV, an unassociated trajectory (Car: 7) is retained for 20 consecutive frames before being deleted. The combined method matches the new detection with the trajectory (Car: 7) solely based on their proximity, resulting in a misassociation (as shown in the second Frame 24), despite the fact that trajectory (Car: 7) should correspond to object (Car: 1237). In contrast, our method, by integrating a deep appearance model, can easily reject such erroneous associations by comparing the similarity of their deep appearance features (as shown in the first Frame 24).

\section{Conclusions}
By combining deep learning and multi-sensor fusion, the proposed MOT algorithm effectively improves tracking performance and robustness.
The proposed MOT algorithm offers several advantages over traditional methods. First, the use of deep learning in appearance extraction allows the algorithm to learn rich and discriminative features that improve object recognition and distinction. Second, the Bi-LSTM-based motion prediction model ensures that the algorithm can adapt to changing motions and maintain accurate tracking over time. Third, the dual-cue (i.e., appearance and motion cues) association strategy ensures that object matches are made not only based on visual similarities but also on their spatial relationships, resulting in more accurate and robust trajectory associations. Finally, the integration of radar-camera sensor fusion enables the algorithm to leverage the complementary information provided by different sensors and address sensor failures, leading to increased accuracy and robustness. The experimental results demonstrate the aforementioned advantages of the proposed method. 
Despite its advantages, current work has limitations. First, the current appearance model, which matches every two objects' appearance, limits operational speed. Second, the current Bi-LSTM motion model lacks a data-driven adaptive update mechanism like Kalman Filtering, preventing it from outperforming Kalman Filters. Future work will focus on resolving these limitations and exploring further enhancements through integrating attention mechanisms with the current Bi-LSTM approach.
%%%%%%%%%%%%%%%%%%%%%%%%%%%%%%%%%%%%%%%%%%%%%

\vspace{0.5cm}

\section*{Acknowledgments}
This work was supported by the Sony Research Award Program.

% \begin{thebibliography} environment in LaTeX is used to manually create a bibliography without relying on a .bib file, and it should not be combined with the \bibliography and \bibliographystyle commands.
\bibliographystyle{IEEEtran}
{\small
\bibliography{references.bib}
}

%\end{thebibliography}

\vfill

\end{document}